\documentclass[journal]{IEEEtran}

\usepackage{amsmath,amsfonts}
\usepackage{bm}
\usepackage{textcomp}

\usepackage{algorithmic}
\usepackage{algorithm}

\usepackage{array}
\usepackage{booktabs} % 三线表
\usepackage{multirow}
\usepackage{makecell}
\usepackage{url}
\usepackage{verbatim}
\usepackage{cite}

\usepackage[caption=false,font=normalsize,labelfont=sf,textfont=sf]{subfig}
\usepackage{stfloats}
\usepackage{graphicx}
\usepackage{wrapfig}
\usepackage{placeins}
\usepackage{float}
\usepackage[dvipsnames,table]{xcolor}
\usepackage[table]{xcolor}

\definecolor{mypink}{rgb}{.99,.91,.95}
\definecolor{mycyan}{cmyk}{.3,0,0,0}
\definecolor{myblue}{RGB}{66,133,244}
\definecolor{mygreen}{RGB}{51,168,83}
\definecolor{myyellow}{RGB}{251,188,3}
\definecolor{myred}{RGB}{234,67,53}
\definecolor{mygrey}{RGB}{95,99,104}
\definecolor{mypup}{RGB}{153,0,204}
\begin{document}

\title{HyperMotionX: The Dataset and Benchmark with DiT-Based Pose-Guided Human Image Animation of Complex Motions}

\author{
    Shuolin Xu$^{\dagger 1,2}$, 
    Siming Zheng$^{\dagger 2}$, 
    Ziyi Wang$^{2}$, 
    Jinwei Chen$^{2}$, 
    Huaqi Zhang$^{2}$ \\
    Daquan Zhou$^{3}$,
    Tong-Yee Lee$^{4}$,
    Hongchuan Yu$^{1}$,
    Bo Li$^{2}$,
    Peng-Tao Jiang$^{2*}$,
    \thanks{$^{\dagger}$ Equal contribution.}%
    \thanks{* Corresponding author.}%
    \\
    Bournemouth University $^1$
    vivo BlueImage Lab, vivo Mobile Communication Co., Ltd $^2$\\
    Peking University $^3$
    National Cheng-Kung University $^4$
}

\maketitle

% --- 摘要 ---
\begin{figure*}[ht!]
    \centering
    \includegraphics[width=1.0\textwidth]{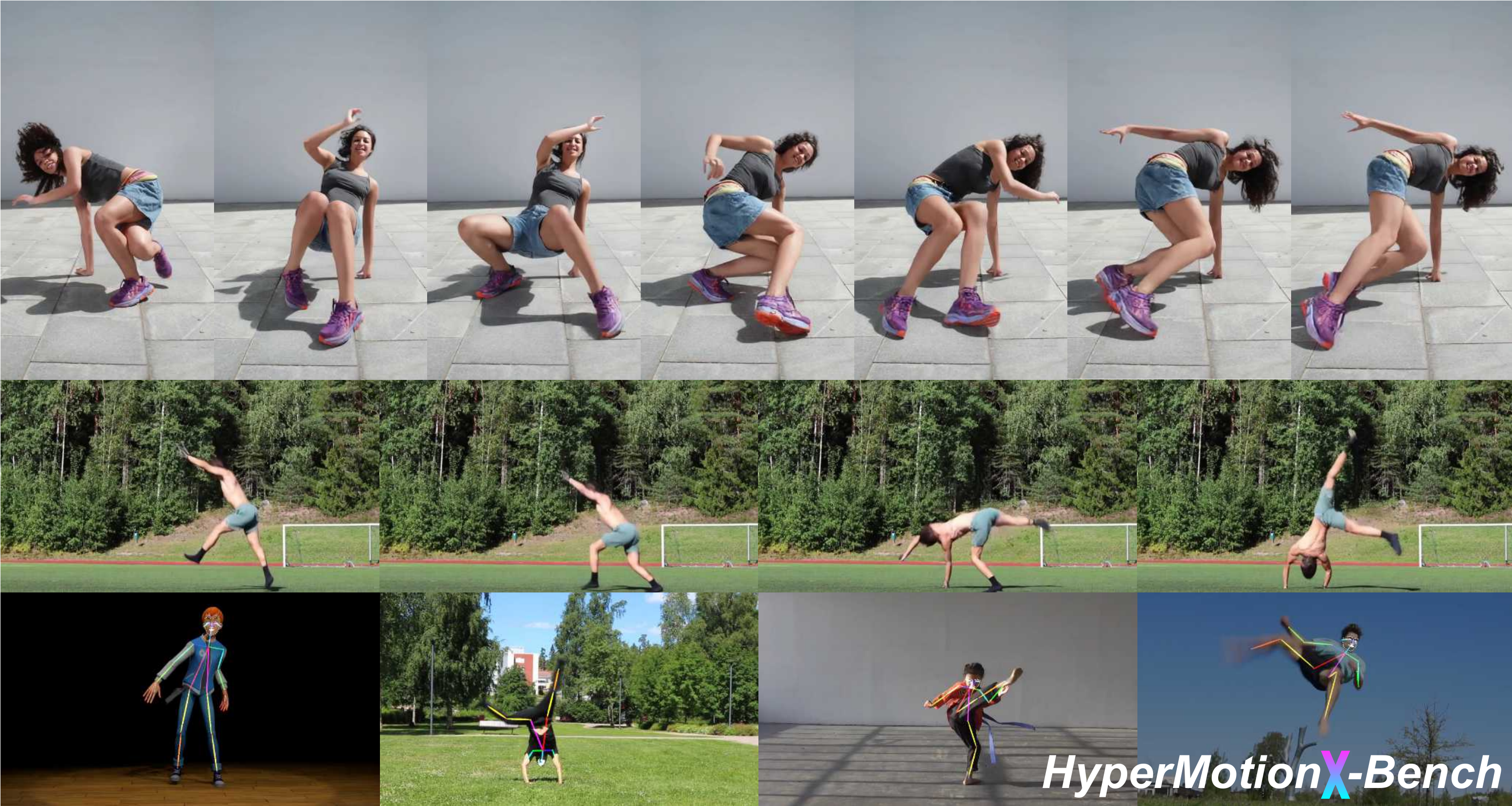}
    \caption{Complex human motion animation samples generated by our method.
We present generation examples in both landscape ($1024 \times 576$) and portrait ($576 \times 1024$) resolutions.}
    \label{fig:teaser}
    \vspace{-0.35cm}
\end{figure*}

\begin{abstract}
Recent advances in diffusion models have significantly improved conditional video generation, particularly in the pose-guided human image animation task. Although existing methods are capable of generating high-fidelity and time-consistent animation sequences in regular motions and static scenes. However there are still obvious limitations when facing complex human body motions that contain highly dynamic, non-standard motions, and the lack of a high-quality benchmark for evaluation of complex human motion animations. To address this challenge, we propose a concise yet powerful DiT-based human animation generation baseline and design spatial low-frequency enhanced RoPE, a novel module that selectively enhances low-frequency spatial feature modeling by introducing learnable frequency scaling. Furthermore, we introduce the \textbf{Open-HyperMotionX Dataset} and \textbf{HyperMotionX Bench}, which provide high-quality human pose annotations and curated video clips for evaluating and improving pose-guided human image animation models under complex human motion conditions. Our method significantly improves structural stability and appearance consistency in highly dynamic human motion sequences.  Extensive experiments demonstrate the effectiveness of our dataset and proposed approach in advancing the generation quality of complex human motion image animations. The codes, model weights, and dataset have been made publicly available at \url{https://vivocameraresearch.github.io/hypermotion/}.
\end{abstract}

 % ===============================================================
%视频生成任务（生成式ai)技术在近期取得了巨大的进展，包括卡通动画领域。但是许多关于真实场景下表现出色的生成任务在卡通领域的表现上却比较差，why；（许多研究者指出【】）当前开源的被用于训练的数据通常是在真实世界情况且设计的模型也是专门为解决现实物理世界的问题，缺乏对卡通动画的关注。而真实世界和卡通动画之间又有巨大的差异，因为卡通动画通常是抽象的夸张的非物理的。生成高质量、可控的、可编辑、（一致性）、包含复杂角色运动的卡通动画视频需要大量的、高质量、多模态、风格多样性的动画数据集。目前此类数据极其稀缺。为了弥补这一差距我们提出了HQMM-CA数据集，这是一个被深度处理和清晰且包含（整体画面、面部画面、面部对照音频、肢体关键点标注、支持多种任务的多模态大规模数据集），专为卡通动画、卡通角色动画生成而设计，以促进社区开发和测试 鲁棒且多功能的动画模型。我们详细介绍了数据集的收集、标注过程及其多模态特征和数据分布。此外我们做了一系列多任务测试，通过结果分析来佐证卡通与现实之间的差距，并在先进的模型上进行微调以显示数据集的价值。
 % ===============================================================

\begin{IEEEkeywords}
Pose-Guided Human image animation, complex motion, datatset and benchmark.
\end{IEEEkeywords}

% --- 正文章节 ---
\section{Introduction}
With the rapid advancement of diffusion models, pose-guided human image animation\cite{hu2024animate, moore2024animateanyone, zhang2024mimicmotion, zhu2024champ, peng2024controlnext, tan2024animate} have achieved remarkable progress. This task focuses on generating temporally coherent human image sequences, conditioned on a reference image of the target person and a corresponding sequence of pose guidance. Recent approaches\cite{hu2024animate,  jiang2023rtmposerealtimemultipersonpose} that integrate human keypoint estimation, high-fidelity image reconstruction, and temporal modeling have demonstrated strong capabilities in synthesizing realistic animations under static scenes or routine actions. These techniques have been widely applied in virtual avatars, motion transfer, post-production, and other creative applications\cite{hu2024animate}.

While existing methods \cite{hu2024animate, zhang2024mimicmotion, peng2024controlnext, tan2024animate} have demonstrated promising performance in general cases, they still struggle to accurately reconstruct human image animation involving complex human motion dynamics, especially for \textbf{Hypermotion:} stunt actions, tricking motions, and acrobatic movements, defined as complex human actions with rapid and atypical motion dynamics.

\begin{figure}[H]
    \centering
    \includegraphics[width=\columnwidth]{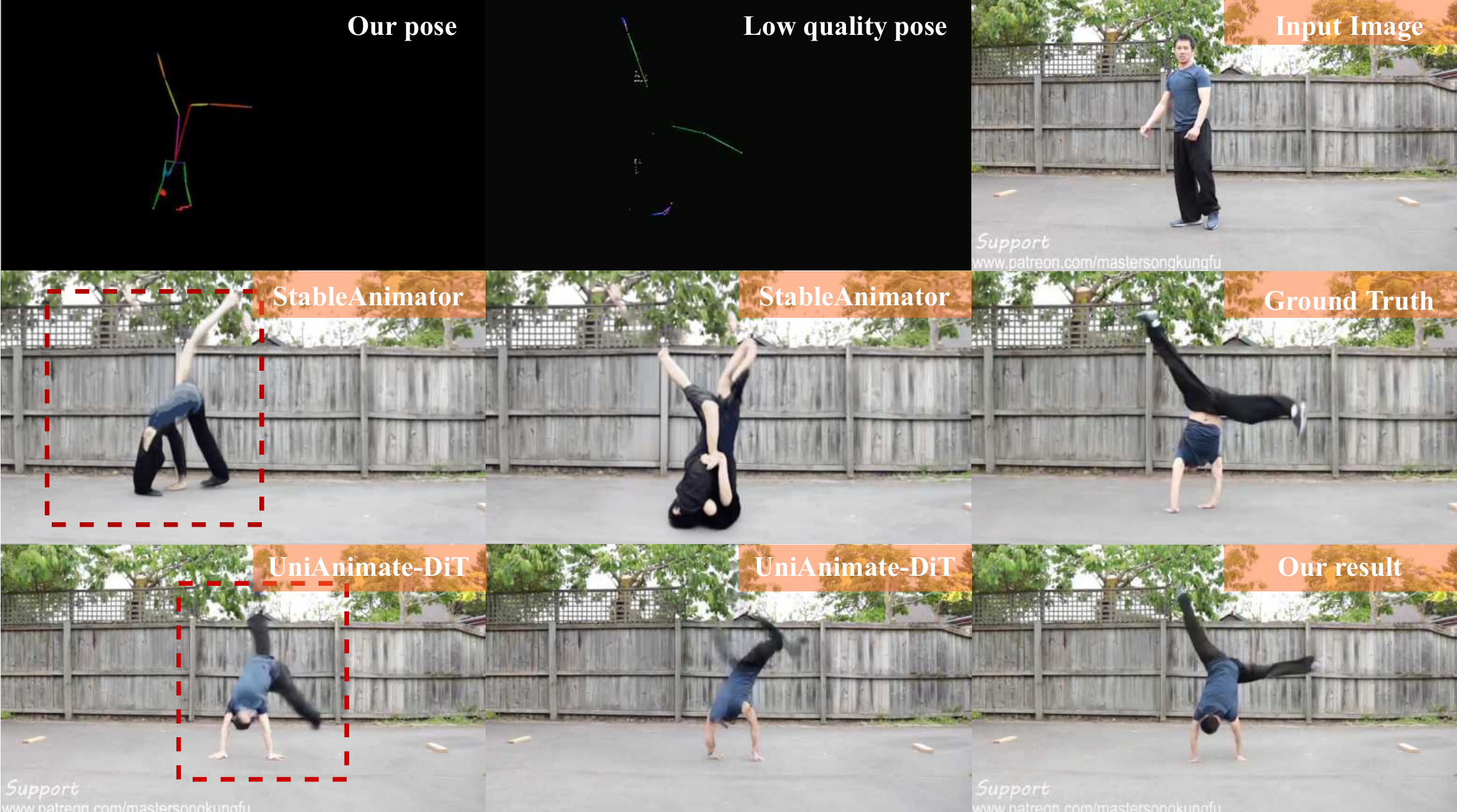}
    \caption{\small Sample video frames of previous methods. When we provide separate high-quality and low-quality pose.}
    \label{fig:motivation}
    \vspace{-0.35cm}
\end{figure}

Existing methods typically rely on external human pose estimation methods such as DWpose\cite{yang2023effective}, Openpose\cite{8765346}, RTMpose\cite{jiang2023rtmposerealtimemultipersonpose} to extract driving pose sequences for animation generation. However, these approaches often fail to accurately capture pose sequences when applied to human videos involving complex motions, as illustrated in Fig.~\ref{fig:motivation}, resulting in a lack of high-quality pose guidance to generate realistic human animation in such scenarios. To address this gap, we propose a new \textbf{Open-HyperMotionX Dataset} and \textbf{HyperMotionX Bench}, centered around complex human motion video clips. The dataset is constructed using our proposed complex human motion region extraction pipeline and is enriched with high-quality human pose annotations for both model training and evaluation.
Our first aim is to verify: \emph{If we provide high-fidelity complex motion pose sequences, can existing human animation models successfully reconstruct human animations with complex motion dynamics?}

We have tested and observed that current mainstream approaches, even when accurate and high-quality pose sequences are provided as guidance, the generated image sequences often suffer from appearance and structural distortions, pose misalignment and pixel region disorder, which corresponds to the low-frequency features severe degradation of the human appearance in the video frames, rather than the texture, edge detail represented by the high frequencies. These issues become particularly prominent in unconventional motion frames (roll, flip, handspring, round kick). The root cause lies in the insufficient spatial modeling and understanding capacity of current models for these non-standard, nonlinear action segments.
These complex motion segments are also characterized by temporal sparsity and are present in only a small fraction of the total video frames. Thus, this temporal sparsity poses a challenge for the model to effectively learn the human body spatial representations in complex motion frames\cite{singer2025videojamjointappearancemotion}.

To address these issues, we argue that the core limitation stems from the insufficient modeling of low-frequency spatial features, which encode global structures and overall body appearance. Under the condition of having high-quality annotated pose sequences in the training dataset, enhancing the model's ability to capture low-frequency spatial features in complex human motion scenes becomes the key to further improving generation quality.
To this end, we first propose a simple DiT-based baseline for the pose-guided human image animation task. It enables the generation of human image animation with large-scale subject motion in open scenes, conditioned on a given reference image and pose sequence.
We then introduce Spatial Low-Frequency Enhancement Rotary Position Embedding (SLF-RoPE), a simple yet effective modification of the standard RoPE mechanism.  
SLF-RoPE selectively amplifies the lowest-frequency channels of the height and width within the frequency tensor by introducing two learnable scaling factors. 
This additional guidance improves the attention mechanism's ability to preserve human body structure, appearance, and spatial consistency, particularly under complex motion conditions, where conventional methods often suffer from structural degradation. 
Furthermore, considering the complex motion frames usually occupy a small portion of the video, and that lengthy video clips are not conducive to model training, we apply Wavelet transform to extract the optimal temporal window through an energy-based optimization algorithm.
Experiments and qualitative results have verified the effectiveness of our method in addressing the complex motion generation.\\
In summary, we make the following contributions:
\begin{itemize}
    \item We propose a simple DiT-based baseline for the pose-guided human image animation task, capable of handling open-world scenarios and large-scale subject motions.    
    \item We introduce Spatial Low-Frequency Enhanced Rotary Positional Embedding, a novel method to selectively enhance spatial low-frequency components, improving appearance and structural fidelity under complex human motions.  
    \item We contribute Open-HyperMotionX Dataset and HyperMotionX Bench, providing high-quality human pose annotations and a benchmark to evaluate pose-guided human image animation models under complex human motion.
\end{itemize}

\section{Related Work}
\subsection{Diffusion for Video Generation}
Recent advances in video generation have been largely driven by diffusion-based frameworks, many of which are adapted from the Stable Diffusion architecture\cite{rombach2022high}. Early methods modify the UNet\cite{ronneberger2015u} to incorporate temporal modeling. 
For example, VDM\cite{ho2022video} extends 2D U-Net to 3D, while Animatediff\cite{guo2023animatediff} integrates 1D temporal attention into 2D spatial blocks for efficiency. 
More recently, transformer-based diffusion models such as DiT\cite{peebles2023scalable} have shown superior performance in visual generation\cite{chen2023pixart}. 
These architectures have been adapted to video tasks\cite{hong2022cogvideo}, yielding variants that either use cross-attention for text embeddings\cite{peebles2023scalable} or jointly attend to concatenated text and visual features\cite{esser2024scaling}. 
In terms of autoencoding, early models relied on standard VAEs\cite{kingma2013auto, rombach2022high}, while recent works such as Hunyuan-Video\cite{kong2024hunyuanvideo} and Wan2.1\cite{wan2025} adopt 3D VAE architectures for improved compression and reconstruction. For textual conditioning, most recent methods employ the T5 family\cite{raffel2020exploring} alongside CLIP\cite{radford2021learning}. Additionally, Rotary Position Embedding (RoPE)\cite{su2024roformer} has become a widely adopted technique for encoding positional information in diffusion transformer models. 

\begin{figure*}[t!]
    \centering
    \includegraphics[width=1\textwidth]{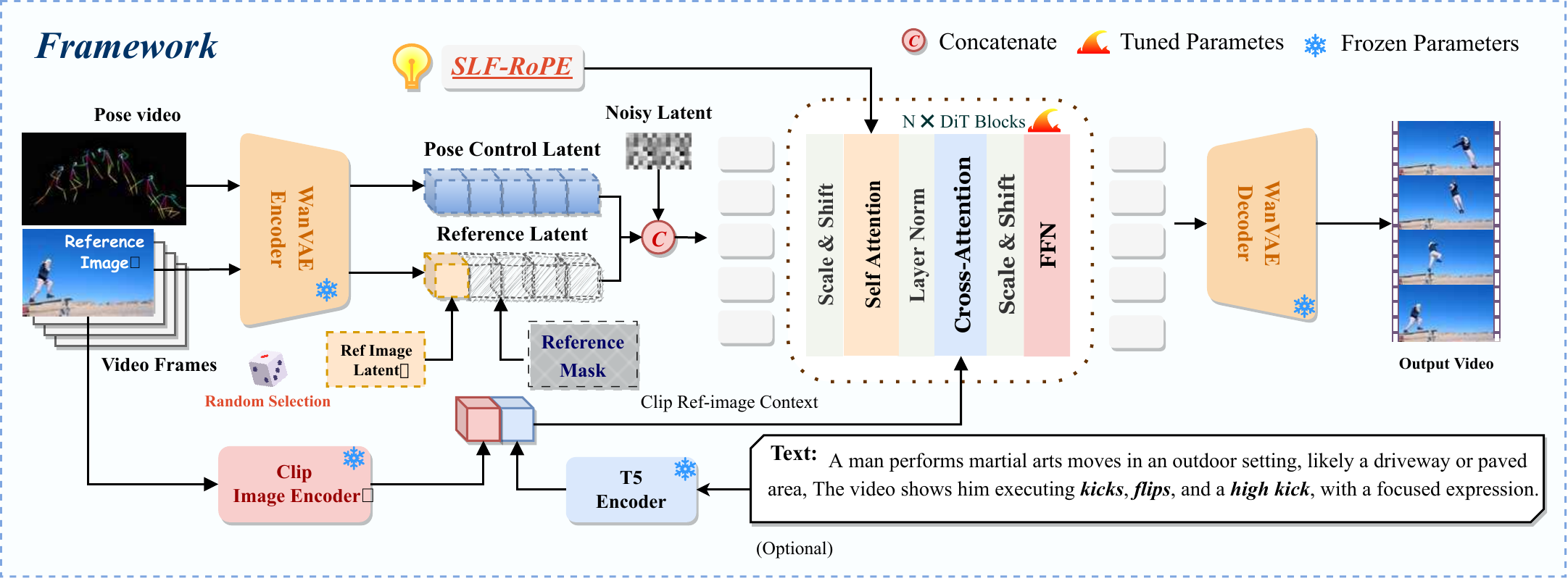}
    \caption{\small
\textbf{The overview of our Hypermotion framework.}
The model takes a reference image and a driving pose video as inputs and generates human animation. Pose control and reference image are injected via latent composition and guided by a binary mask. Spatial Low-Frequency Enhanced RoPE is applied in self-attention.}
    \label{fig:pipeline}
    \vspace{-0.35cm}
\end{figure*}

\subsection{Pose-guided Human Image Animation}
Pose-guided human animation generation is the task of synthesizing temporally coherent, photorealistic human videos whose appearance matches a given reference human image identity while following an input target pose sequence\cite{hu2024animate, zhang2024mimicmotion}. Early works\cite{li2019dense, siarohin2019first, siarohin2021motion} based on Generative Adversarial Networks (GANs)\cite{goodfellow2020generative} to generate animation from source image. However, these GANs-based models often show various artifacts in the generated animation results. 
Recently, animation based on the diffusion models\cite{wang2024disco} emerged 
and achieved pleasing human animation results. 
\cite{hu2024animate, xu2024magicanimate} both include a pose net to process pose information and a reference net to model appearance, with the introduction of the temporal attention layer\cite{guo2023animatediff} enhancing temporal consistency.
In addition, more representation information has been introduced such as depth and 3D signal SMPL to enhance the controllable capability, as represented by Champ\cite{zhu2024champ, zhou2024realisdance}. ControlNeXt\cite{peng2024controlnext} and Mimicmotion\cite{zhang2024mimicmotion} all utlize post-processing to deal with facial distortion. Stableanimator\cite{tu2024stableanimator} introduced the face encoder and combined it with the face mask to address this problem. 
Animate-X\cite{tan2024animate} has developed the Pose Indicator to generalize the model to the animation of anthropomorphic characters. 
Follow-Your-Pose V2\cite{xue2024followv2} enhanced implicit decoupling towards multi-character image animation. 
UniAnimate\cite{wang2025UniAnimate} introduced an additional Mamba module. HumanVid\cite{DBLP:conf/nips/00010ZF0LTCX0L24} added the camera position parameter to the model, allowing the generation of animations that incorporate camera motion and provided a human image animation dataset. 
With the recent success of the video diffusion transformer\cite{peebles2023scalable} model, UniAnimate-DiT\cite{wang2025UniAnimate} , DreamActor\cite{luo2025dreamactor}, have started to extend the task to a DiT-based model and have demonstrated high-quality generation performance, %MTVcrafter\cite{ding2025mtvcrafter4dmotiontokenization} used 4D motion tokenization to improved the task.
Despite promising results, current methods struggle on complex motions; even with high-quality pose inputs, outputs show structural artifacts and misalign with the driving poses.
%Although these methods have achieved promising results, they still struggle to generate realistic animations for complex human motions. Even when provided with high-quality pose sequences as driving input, the generated human figures often suffer from body structure anomalies and fail to align accurately with the driving poses.
\section{Methodogy}
\subsection{Preliminaries}
\textbf{Diffusion Transformer (DiT).}
The DiT\cite{peebles2023scalable} proposes a novel design that combines the generative strengths of diffusion models with the representational power of transformer architectures\cite{vaswani2017attention}.
This combination effectively addresses the inherent limitations of traditional UNet-based latent diffusion models (LDMs), resulting in improvements in generation quality, model flexibility, and scalability. Therefore, we propose using DiT as the base model to enhance the quality of human animation generation for complex motions.

\textbf{Rotary Position Embedding (RoPE).}
Rotary Positional Embedding (RoPE)\cite{su2024roformer} has become the standard way of injecting position information into modern Transformer layers.
For a 1-D token sequence, let
$\bm{x}\in\mathbb{R}^{d}$ denote the input at position
$p\in\mathbb{N}$.  We select the first $d'\!\le d$ channels
(assumed even) and apply a position-dependent rotation to each channel
pair:
\begin{equation}
\label{eqn:RoPE-Complex}
\boldsymbol{f}^{\text{RoPE}}(\bm{x}, p, \bm{\theta})_j = 
   \begin{bmatrix}
       \cos(p\,\theta_j) & -\sin(p\,\theta_j) \\
       \sin(p\,\theta_j) & \cos(p\,\theta_j)
   \end{bmatrix}
   \begin{bmatrix}
       x_{2j} \\
       x_{2j+1}
   \end{bmatrix},
\end{equation}
\noindent
where $\bm{\theta} \in \mathbb{R}^{d'/2}$, with the frequency components defined as
$\theta_j = b^{-2(j-1)/d'}$ for $j = 1, \ldots, d'/2$.  
The vector $\bm{\theta}$ specifies the positional encoding frequencies across all rotated channel dimensions, and $b$ serves as a scaling hyperparameter controlling the base frequency.  
The inner product between two RoPE-embedded vectors is dependent solely on their relative positional offset, enabling relative position encoding without introducing additional learnable parameters.  
In practical implementations, RoPE is applied to both query and key vectors prior to the dot-product operation within the attention mechanism, such that the resulting attention weights naturally encode relative positional information.

\textbf{Notations.}
Let $\bm{L}_{\text{noisy}} \in \mathbb{R}^{B \times C \times T \times H \times W}$ denote the noisy latent video sequence, where $B$ is the batch size, $C$ is the channel dimension, $T$ is the frame length, and $H, W$ represent spatial height and width.  
The objective of Latent Video Diffusion Models (LVDM) is to progressively denoise $\bm{L}_{\text{noisy}}$ to produce realistic video samples in the latent space before decoding back to pixel space.
\begin{algorithm*}[t]
\caption{SLF-RoPE Frequency Scaling}
\label{alg:slfrope_twocol}
\begin{algorithmic}[1]
\REQUIRE Frequency tensor $\theta$ decomposed into temporal ($t$), height ($h$), and width ($w$) components
\REQUIRE Learnable parameters \textit{motion\_scale} and \textit{space\_scale\_factor}
\REQUIRE Low-frequency ratio $\alpha = 30\%$
\STATE Compute scaling factor $\gamma \leftarrow 1 + \textit{space\_scale\_factor} \cdot \textit{motion\_scale}$
\FOR{each spatial axis $a \in \{h, w\}$}
    \STATE Identify low-frequency indices:
    $I_{\text{low}} \leftarrow \text{last } \alpha\% \text{ of } \theta^{(a)}$
    \STATE Scale low-frequency components:
    $\theta^{(a)}[I_{\text{low}}] \leftarrow \gamma \cdot \theta^{(a)}[I_{\text{low}}]$
\ENDFOR
\STATE Apply the modified $\theta$ in the standard RoPE rotation
\end{algorithmic}
\end{algorithm*}

\subsection{Simple DiT-based baseline}
Building upon the Wan2.1\cite{wan2025} and Wan2.1-Fun-Control I2V\cite{Bub2025videoX} video generation models, we propose a DiT-based baseline for the pose-guided human image animation task with minimal architectural modifications, as illustrated in Fig.~\ref{fig:pipeline}. Unlike prior human image animation baselines, our method does not require additional reference networks\cite{hu2024animate}, ControlNet\cite{zhang2023adding} modules, or insertion of reference tokens into the token sequence with corresponding adjustments to positional embeddings or Rotary Position Embedding (RoPE)\cite{su2023roformerenhancedtransformerrotary}.  
Furthermore, it eliminates the need for a separately trained PoseNet\cite{hu2024animate}. In contrast, PoseNet based approaches introduce additional parameters and require extra feature alignment with the backbone after extracting pose video representations, which may inevitably incur information loss. Despite its simplicity, the baseline can generate human animations across diverse, open-domain scenes by conditioning on a reference image and a driving pose sequence.

The Wan2.1 I2V baseline\cite{wan2025} introduces a conditional image as the first frame and applies a Wan-VAE compression to obtain latent guidance features.  
The condition image $I \in \mathbb{R}^{C \times 1 \times H \times W}$ is concatenated with zero-filled frames to form $I_c$, then compressed to latent $z_c$.  
A binary mask $M$ marks the preserved frame for conditioning.  
Latent noise $z_t$, condition $z_c$, and the mask $m$ are concatenated and passed to the DiT backbone for generation.  
The reference image is encoded via CLIP\cite{radford2021learning}, combined with text tokens from umT5\cite{chung2023unimax}, and injected through the Decoupled cross-attention. 

\subsection{Latent Composition for Conditional Control}
We used a unified latent space composition framework to inject both pose and reference image conditions into the DiT-based video generation pipeline.

\textbf{Pose Control Latent Inject.}
To enable the pose sequence to serve as a condition for guiding animation, our baseline eliminates the need for an additional 3D convolution-based pose network requiring extra training. Instead, the input pose video $I_{\text{pose}}$ is directly processed by the Wan-VAE to obtain a pose control latent $\mathbf{L}_{\text{pose}} \in \mathbb{R}^{B \times 16 \times T \times H \times W}$ that matches the shape of the noise-added latent $\mathbf{L}_{\text{noisy}} \in \mathbb{R}^{B \times 16 \times T \times H \times W}$ features. The resulting pose control latent is then concatenated with the latents along the channel dimension $\mathcal{C}$ before being fed into the denoising network.

\textbf{Reference Image Inject.}
The key challenge is how to inject the reference image such that any provided pose sequence can consistently drive the target appearance. The Wan2.1\cite{wan2025} I2V base model does not support this capability and often suffers from misalignment between the input image and the driving pose, resulting in identity inconsistency and frame artifacts. Thus, the reference image should not be concatenated directly with the noisy latent as in the original I2V baseline.

To address this, we propose a simple yet effective mechanism based on latent space composition and masking.  
If a reference image is provided, it is preprocessed and encoded by Wan-VAE to obtain a reference latent $\mathbf{L}_{\text{ref}} \in \mathbb{R}^{B \times 16 \times 1 \times H \times W}$.  
A zero-initialized tensor with the same shape as the noisy video latents $\mathbf{L}_{\text{noisy}} \in \mathbb{R}^{B \times 16 \times T \times H \times W}$ is created, and $\mathbf{L}_{\text{ref}}$ is inserted into the first frame (index 0). %
Meanwhile, a binary reference mask $\mathbf{M}_{\text{ref}} \in \{0, 1\}^{B \times 4 \times T \times H \times W}$ is constructed, where only the first frame is set to 1.0 and all others to 0.  
This mask explicitly controls the spatio-temporal influence of the reference image, ensuring that its appearance only guides the first frame and prevents leakage into subsequent frames.
Finally, the reference latent tensor and the reference mask are concatenated with the pose control latents $\mathbf{L}_{\text{pose}}$ and the noisy video latent $\mathbf{L}_{\text{noisy}}$ along the channel dimension to form the final input for the denoising network:
\begin{equation}
\bm{L}_{\text{final}} = \text{Concat}(\mathbf{L}_{\text{noisy}},\ \mathbf{L}_{\text{pose}},\ \mathbf{L}_{\text{ref}},\ \mathbf{M}_{\text{ref}}) \in \mathbb{R}.
\end{equation}

\textbf{Spatial Low-Frequency Enhanced RoPE (SLF-RoPE).} The SLF-RoPE is a concise yet effective modification to the frequency formulation of RoPE that selectively amplifies the lowest frequency channels of the height and width axes.  
Our intuition is that low-frequency spatial components encode large-scale shapes, structures, and basic appearance layouts, which are most critical for maintaining global consistency under fast motions.

Following the standard RoPE definition, the frequency vector $\bm{\theta}\in\mathbb{R}^{d'/2}$ is computed as:
\begin{equation}
\bm{\theta}_j = b^{-2(j-1)/d'}, \quad j = 1, \ldots, d'/2,
\label{eq:rope-freq}
\end{equation}
where $b$ is a base scaling factor.
In SLF-RoPE, we partition $\bm{\theta}$ into temporal ($t$), height ($h$), and width ($w$) frequencies.  
For the spatial axes, we define the low-frequency range as the bottom $\alpha$ fraction of channels.  
An extra scaling factor $\gamma > 1$ is then applied to these low-frequency components:
\begin{equation}
\bm{\theta}^{(h)}_{\text{low}} = \gamma \cdot \bm{\theta}^{(h)}_{\text{low}}, \quad
\bm{\theta}^{(w)}_{\text{low}} = \gamma \cdot \bm{\theta}^{(w)}_{\text{low}}.
\label{eq:slf-rope}
\end{equation}

In our design, $\alpha$ is set to $30\%$, and $\gamma$ is dynamically modulated by two learnable parameters: the motion scale and the space scale factor.  
The motion scale controls the scaling strength based on the estimated motion intensity of the input video, while the space scale factor controls the global amplification ratio of the spatial low-frequency bands.  
The modified frequency tensor is then applied identically as in standard RoPE during complex-valued rotation in the attention mechanism.  
This dynamic enhancement allows self-attention layers to better preserve spatial layouts and structural consistency in highly dynamic motion regions, demonstrated in the experimental section\ref{subsec: ex}. Full pseudocode is given in~\ref{alg:slfrope_twocol}.

\begin{figure*}[t]
  \centering
  \includegraphics[width=0.85\textwidth]{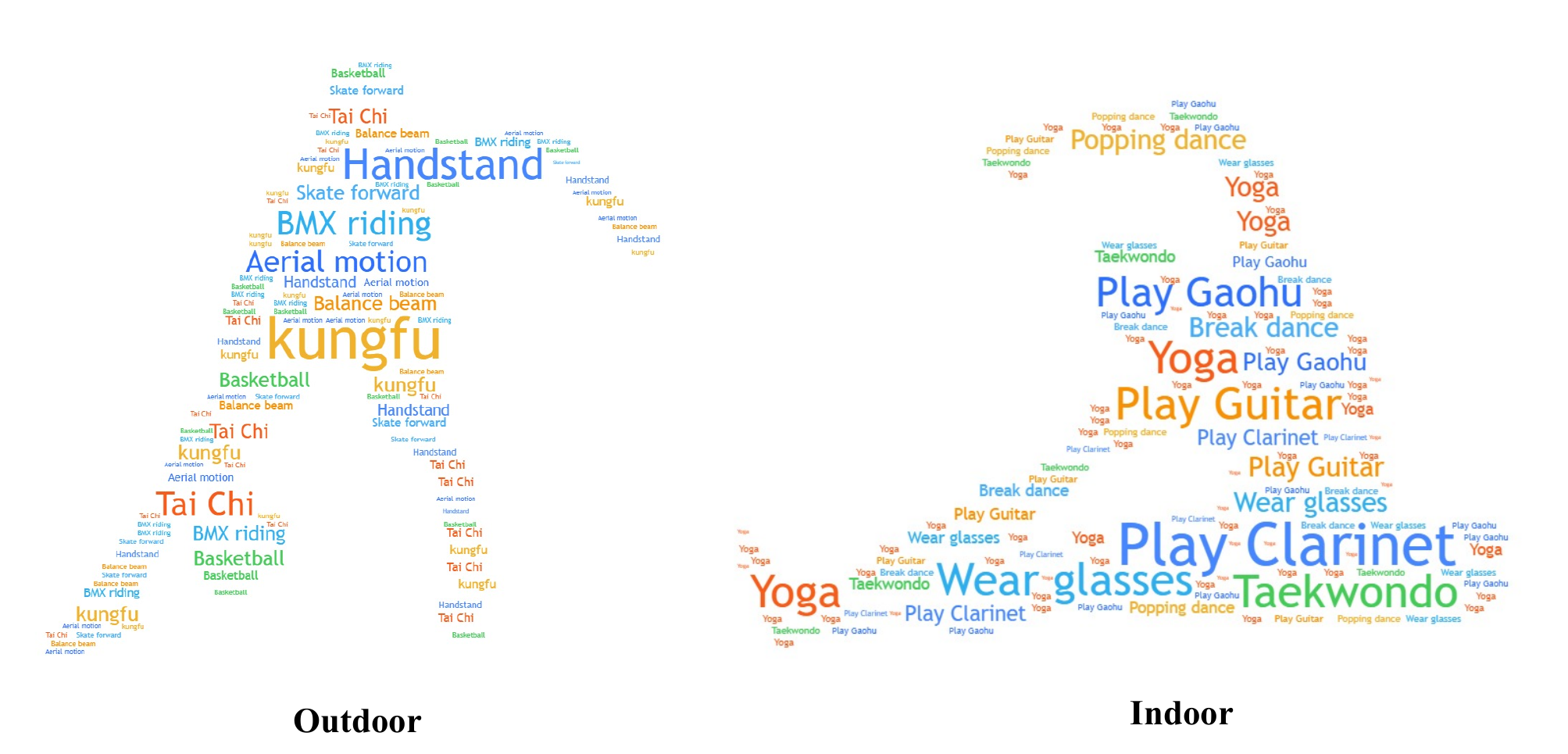}
  \caption{\textbf{Motion categories in Open-HyperMotionX.}
  The dataset encompasses diverse indoor and outdoor human motion categories of varying complexity.}
  \label{fig:cloud}
\end{figure*}

\subsection{Open-HyperMotionX Dataset}
\label{dppp}
At present, there is no publicly available dataset specifically designed for the generation and evaluation of complex human motion videos with accurate pose sequence annotations. To address this gap, we propose a new dataset, the \textbf{Open-HyperMotionX}. We curate source videos from the open-source MotionX\cite{zhang2025motion, lin2023motionx} dataset, which was originally developed for tasks such as digital humans, motion capture, motion generation, and human mesh recovery. MotionX\cite{zhang2025motion, lin2023motionx} provides a large collection of human motion videos, including martial arts, gymnastics, stunt actions, and other highly dynamic performances, typically captured with a fixed camera and continuous long takes. However, the dataset has not been directly associated with video generation tasks before.

Through a dedicated data processing pipeline and additional annotations, we refine and extend the video and 2D keypoint annotations of MotionX\cite{zhang2025motion, lin2023motionx} to create Open-HyperMotionX, tailored for the study of complex motion human image animation.

The \textbf{Open-HyperMotionX} dataset contains 19,597 video clips comprising 2,507,811 frames, with a total duration of approximately 26 hours and 40 minutes. The average duration of a single clip is 5 seconds, with the majority ranging between 5 and 6 seconds. The dataset has an average frame rate of 26.4 fps, with most videos recorded at either 25 or 30 fps. In terms of spatial resolution, the dominant format is 1920$\times$1080 or 1080$\times$1920, which accounts for the majority of the clips. 

We have provided a detailed comparison table of statistical metrics between our collected dataset and existing publically available datasets: TikTok dataset\cite{Jafarian_2021_CVPR}, UBC-fashion\cite{Zablotskaia2019DwNet}, HumanVid\cite{DBLP:conf/nips/00010ZF0LTCX0L24}(As many of the official links provided are no longer available, we were only able to download 10,136 videos.) and our base dataset motionX\cite{lin2023motionx}. The Table~\ref{tab:data_compare} includes clip counts, average durations, resolutions, these statistics clearly highlight the advantages of our dataset over existing datasets in terms of data volume, resolution, video specification and duration distribution.

\begin{table*}[t]
    \centering
    \caption{Comparison of our Open-HyperMotion dataset with existing datasets.}
    \label{tab:data_compare}
    \small
    \setlength{\tabcolsep}{6pt}
    \renewcommand{\arraystretch}{1.15}
    \begin{tabular}{lccccp{3.2cm}p{2.0cm}c}
        \toprule
        \textbf{Dataset} &
        \textbf{Clips} &
        \textbf{Total Frames} &
        \textbf{Avg. Frames} &
        \textbf{Avg. Duration} &
        \textbf{Resolution} &
        \textbf{Total Duration} &
        \textbf{FPS} \\
        \midrule
        TikTok-dataset
        & 340
        & 92,961
        & 273
        & 9s
        & 604$\times$1080
        & 3,098s
        & 30 \\
        
        UBC-Fashion
        & 600
        & 231,373
        & 385
        & 12.8s
        & 720$\times$940
        & 7,800s
        & 30 \\
        
        HumanVid
        & 10,136
        & 4,533,721
        & 447
        & 17.6s
        & 1080$\times$1920; 1440$\times$2732
        & 182,448
        & 25/30 \\
        
        MotionX
        & 25,859
        & 4,428,694
        & 171
        & 7s
        & 1920$\times$1080; 1280$\times$720; 852$\times$480; 640$\times$360
        & 163,567s
        & 25/30 \\
        
        \midrule
        \rowcolor{gray!10}
        \textbf{HyperMotionX}
        & \textbf{19,597}
        & \textbf{2,507,811}
        & \textbf{128}
        & \textbf{5.2s}
        & \textbf{1920$\times$1080; 1280$\times$720}
        & \textbf{96,040s}
        & \textbf{25/30} \\
        \bottomrule
    \end{tabular}
\end{table*}

\textbf{Data Process.}
For the pose-guided human image animation task, the ideal training data should consist of video clips with continuous shots (no cuts), full body visibility, minimal occlusion, and only slight camera movement. The source videos in MotionX\cite{lin2023motionx} largely satisfy these criteria, which is a key reason why we chose to follow and build upon the MotionX\cite{lin2023motionx} dataset. However, additional processing is required for video generation involving complex human motions. We follow the data processing pipeline proposed by EasyAnimate\cite{xu2024easyanimatehighperformancelongvideo}, incorporating Wavelet Transform Energy Analysis into the pipeline specifically for capturing complex motion within temporal regions. Then we performed the following processing steps:
\begin{itemize}
  \item We applied YOLOv8\cite{yolov8_ultralytics} to keep clips with a single person; keypoint trajectories were linearly interpolated and outliers removed to denoise annotations.
  \item To target complex motion regions in long videos, we analyzed 2D keypoints with a \textbf{wavelet transform} and selected optimal windows via an energy-based criterion, capturing representative segments (e.g., the airborne phase of a cartwheel).
  \item We reduced text region negative interference for training by detecting text with EasyOCR\cite{easyocr} and applying Gaussian blur masks.
  \item Following EasyAnimate\cite{xu2024easyanimatehighperformancelongvideo}, captions were first generated in batch by InternVL2.0\cite{chen2024internvl} and then refined with LLaMA-8B\cite{llama3modelcard}, prioritizing action-related keywords while minimizing detail of scene appearance.
\end{itemize}

\subsection{Complex human motions video clip extraction via wavelet transformer}
\label{WAVELET TRANSFORMER}
\begin{figure*}[t]
  \centering
  \includegraphics[width=\textwidth]{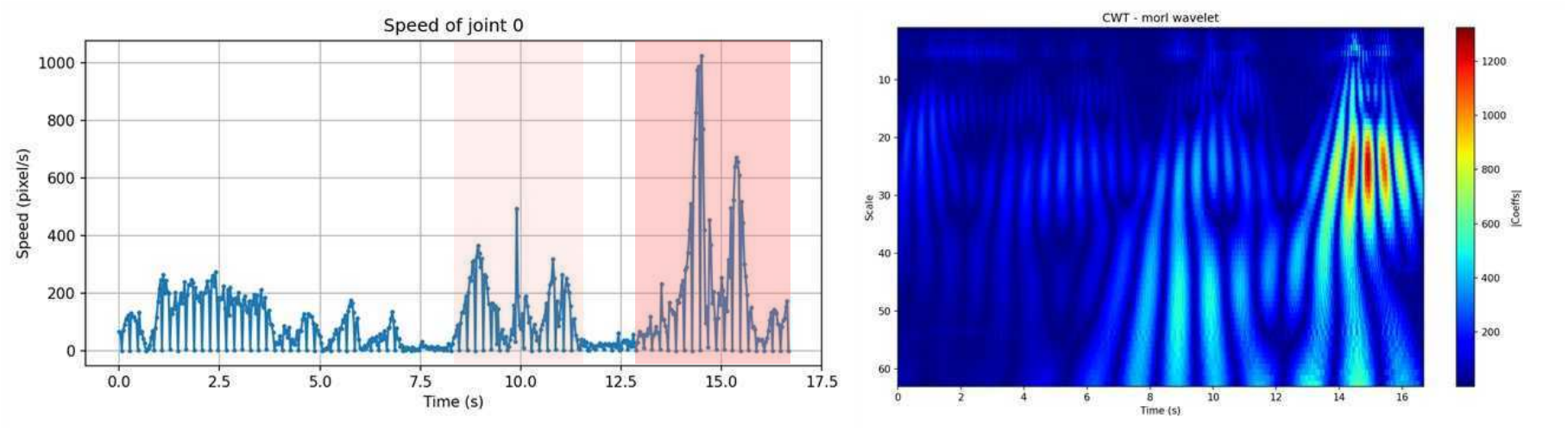}
  \caption{\textbf{Automatic selection of representative motion segments.}
  To extract the most representative action segments from long video sequences, we propose an automated method based on human keypoint motion characteristics. Specifically, we apply wavelet transforms to analyze temporal keypoint motion patterns and formulate an energy-based optimization strategy to select the optimal time window. The figure highlights the extracted complex motion intervals, corresponding to vacating phases during flipping actions.}
  \label{fig:motion-time}
\end{figure*}

We observe that during moments when the human body performs highly challenging and complex motions, the acceleration curves computed from human keypoints exhibit pronounced fluctuations. To extract the most representative action segments from long video sequences, we developed an automated method based on human keypoint motion characteristics. This method utilizes the wavelet transform to analyze human keypoint movement patterns and employs an energy optimization algorithm to select the optimal time window. It can effectively help us capture high quality complex motion segments within lengthy data. As shown in Fig \ref{fig:motion-time}. The specific steps are as follows:

\subsubsection{Motion Velocity Calculation}
Based on the extracted keypoint sequences, we calculate the inter-frame velocity of a specific joint (default index 0, typically corresponding to the head or neck):
\begin{equation}
\label{eqn:fps}
v(t) = \sqrt{(x_{t+1} - x_t)^2 + (y_{t+1} - y_t)^2} \times \text{fps}
.\end{equation}
where $x_t$ and $y_t$ are the coordinates of the joint at frame $t$, and $fps$ is the frame rate of the video. This step converts spatial position changes into time series data, providing a foundation for subsequent analysis.

\subsubsection{Wavelet Transform Energy Analysis}
To capture time-frequency characteristics in the motion, we apply \textbf{Continuous Wavelet Transform (CWT)} to the velocity sequence:
\begin{equation}
\label{eqn:cwt}
CWT(a,b) = \frac{1}{\sqrt{a}} \int_{-\infty}^{\infty} v(t) \psi^*(\frac{t-b}{a})dt
,\end{equation}

where $\psi$ is the Morlet wavelet function, $a$ is the scale parameter, and $b$ is the translation parameter. We calculate wavelet coefficients across multiple scales (1 to 128) and obtain an energy sequence by taking the absolute value of all scale coefficients and summing them:
\begin{equation}
\label{sumcwt}
E(t) = \sum_{a=1}^{a_{max}} |CWT(a,t)|
.\end{equation}
This energy sequence reflects the complexity and intensity of motion at different time points.

\subsubsection{Energy Peak Optimization}
To enhance the robustness of the algorithm, we apply peak width filtering to remove narrow peaks that may be caused by noise (the noise in keypoint velocity curves is primarily introduced by sporadic errors in keypoint localization, often leading to sharp, nonphysical jumps between adjacent frames.):
\begin{equation}
\label{epo}
E'(t) = \begin{cases} 
E(t) & \text{if peak width} \geq \text{threshold} \\
0 & \text{otherwise}
\end{cases}
.\end{equation}
In our implementation, the peak width threshold is set to 3 frames to ensure that only meaningful movements with sufficient duration are retained.

\subsubsection{Optimal Time Window Selection}
Finally, we employ a sliding window approach to find the best segment of fixed length (default 6 seconds) on the processed energy sequence:
\begin{equation}
\label{window}
W_{best} = \arg\max_{i} \sum_{j=i}^{i+L-1} E'(j)
,\end{equation}
where $L$ is the target window length (in frames). To handle boundary cases, when the optimal window is too close to the beginning or end of the video (less than 10 frames away), we adjust the window to start from the beginning or end 6 seconds before the end of the video. For videos shorter than the target length, we retain the original video without cropping. During the early stage of constructing Open-HyperMotionX, We identified numerous cases of missing annotations and noisy data from the original MotionX dataset\cite{lin2023motionx} then manually screened the original data. The automated curation pipeline we proposed was designed precisely in response to these initial observations.

\subsubsection{Video Segment Extraction Implementation}
Based on the above method, we use FFmpeg for precise video cropping, ensuring that the output segments have consistent duration and contain the most representative actions. To ensure video quality, we re-encode the video segments using H.264 encoding (CRF=23).

Compared to traditional fixed time point sampling or manual selection, our method can automatically identify and extract video segments containing the richest motion information, providing more consistent and high-quality input data for subsequent action recognition and analysis.

\subsection{HyperMotionX Bench}
Our initial motivation stems from the current limitation that accurate and high-quality pose sequences for complex human motions cannot be reliably extracted using existing models such as DWpose\cite{yang2023effective}, Openpose\cite{8765346}, RTMpose\cite{jiang2023rtmposerealtimemultipersonpose}. This raises the question: \emph{if more accurate and refined pose sequences were provided to existing methods, could they successfully generate high-quality videos of complex human motions? }Furthermore, due to the lack of a dedicated benchmark for evaluating complex motion pose-guided human image video generation, we introduce the \textbf{HypermotionX Bench} as an additional dataset for our primary task of pose-driven human image animation.

\textbf{HypermotionX Bench} is designed to assess the quality of complex human motion generation while providing more accurate pose annotations and the corresponding pose-guided video sequences. The source videos were manually collected and selected from publicly available websites that meet open source criteria and processed following the same pipeline as Open-HyperMotionX. After testing, we found that Xpose\cite{xpose} extracts the best performance of pose in complex human motion videos, so we use Xpose\cite{xpose} to extract the whole body pose for our bench, due to the low accuracy of hand detection, we discarded the hand annotation for some of the most complex cases frames. After careful manual selection, the final benchmark contains 100 complex human motion video clips, primarily in landscape orientation with resolutions centered around 1080p. The benchmark covers a wide range of complex and basic human actions, with the complex motions mainly sourced from Tricking (martial arts), including classic movements such as Forward Rolls, Backward Rolls, Front Handspring, Back Handspring, Cartwheel, Front Flips, Webster, Back Flips, Moon Kick, Side Flips, and 360 Kick. More types of actions are illustrated in Fig.~\ref{fig:cloud}. Finally, it is important to note that, to ensure fair evaluation, all cases included in the HypermotionX Benchmark were strictly separated from the Open-HyperMotionX dataset used for training.

\section{Experiments}
\label{subsec: ex}
\subsection{Implementation Details}
\label{subsec: imple}
Based on the Wan2.1-Fun-Control\cite{Bub2025videoX} pre-training weights, we conducted training and experiments on the 14B parameters version of the model using our Open-HyperMotionX dataset. We conducted extensive experiments based on the Wan2.1-Fun framework, trained on our proposed Open-HyperMotionX dataset. For the 14B parameter model, we performed full fine-tuning for a total of 30{,}000 steps using the ZeRO Stage 2\cite{rajbhandari2020zero} optimization strategy on 8 NVIDIA H20 GPUs, which required approximately 12 days of training. The learning rate for experiments was uniformly set to $5 \times 10^{-6}$. During inference, we set the Guide scale to 6.0, only a single RTX 3090 GPU is required which consistently produced satisfactory generation quality in various test samples.

\begin{figure*}[t]
    \centering
    \includegraphics[width=\textwidth]{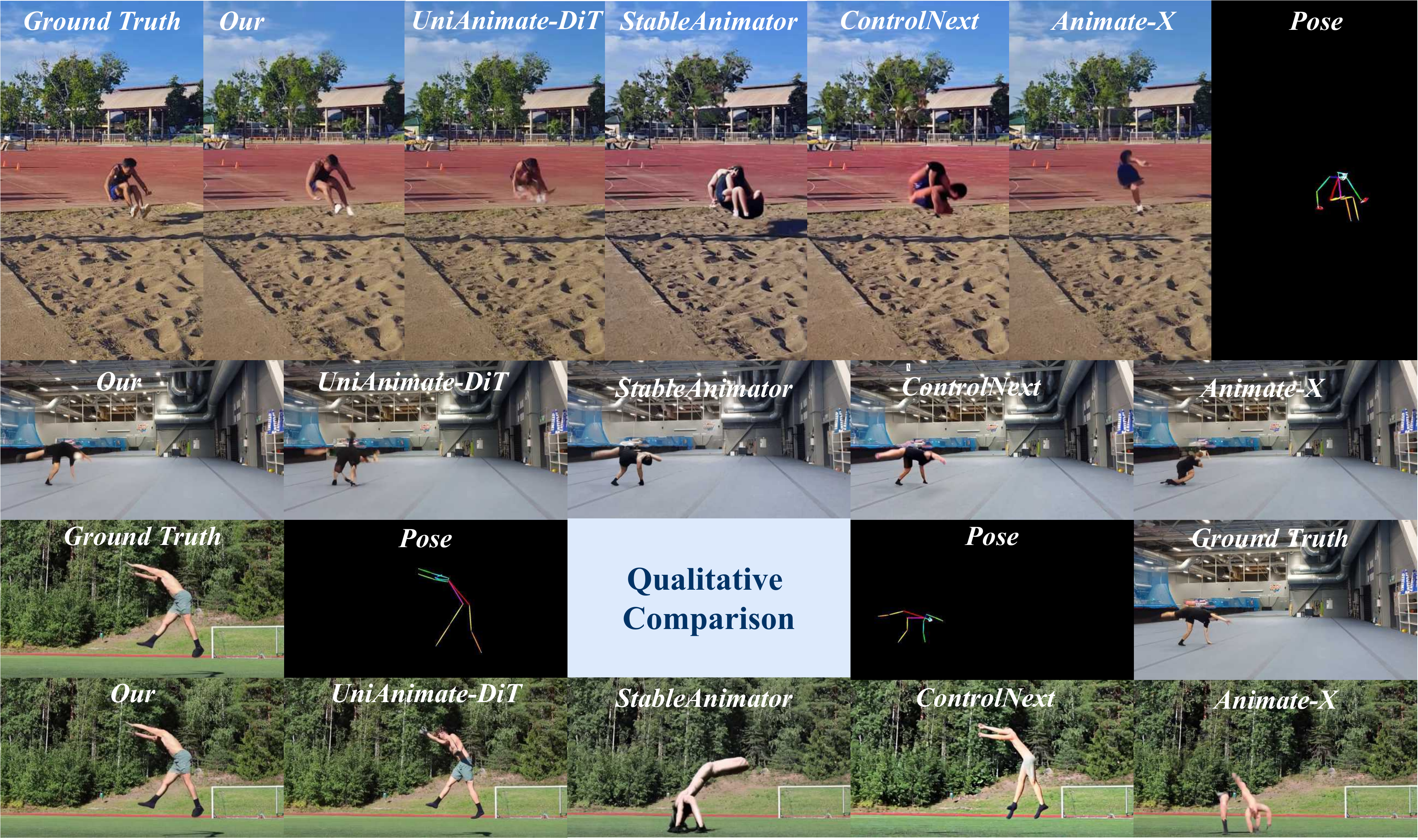}
    \caption{Qualitative comparison with state-of-the-art methods.
    Our method demonstrates superior structural coherence, appearance consistency, and motion stability
    under complex human motions (e.g., front flips).
    Results are shown in both $1024\times576$ (landscape) and $576\times1024$ (portrait) resolutions.}
    \label{fig:D}
\end{figure*}

In implementing \textbf{SLF-RoPE}, we replaced the original Rotary Positional Embedding (RoPE) used in the self-attention modules and introduced two learnable scaling parameters: the Motion Scale and the Space scale factor. The motion scale was initialized to 1.5 and the space scale factor was initialized to 0.02. Additionally, within the \textbf{SLF-RoPE} module, we set the low ratio to 30\%, which determines the proportion of low-frequency channels along the spatial height and width dimensions.

\subsection{Experimental Setup}
As our work focuses on complex human animation and we propose the HypermotionX Bench specifically to evaluate this task, we provide not only the dataset, but also high-quality pose sequences and a reproducible method to extract these poses. To ensure fairness in evaluation, we made every effort to allow all comparison methods to use the same pose annotation provided by our benchmark.

Since there is no unified standard for frame rate, frame count, or resolution across different baseline methods, we carefully adapted the input settings of each method to be compatible with the HypermotionX Bench while preserving the integrity of a fair comparison. Specifically, for ControlNeXt\cite{peng2024controlnext,tu2024stableanimator}, we used the pre-processed pose videos provided by us and adjusted the resolution, frame rate, and frame count accordingly. Similarly, for UniAnimate-DiT\cite{wang2025UniAnimate}, the resolution was set to match our preprocessed pose videos; however, due to its fixed frame count requirement, we dynamically adjusted the frame rate to align with the duration of the pose video provided. For Animate-X\cite{tan2024animate}, since the frame rate is fixed at 8 fps, the total frame number was calculated directly based on the duration of the corresponding pose video.

We conducted a comprehensive quantitative evaluation of our method and recent state-of-the-art baselines on the proposed HyperMotionX Bench.
Additionally, we evaluated the effectiveness of our dataset for training pose-driven human image animation models.
We adopted widely used metrics including Pixel fidelity: PSNR\cite{hore2010image}, SSIM\cite{wang2004image}, L1-Loss and Perceptual quality: FID\cite{heusel2017gans}, LPIPS\cite{zhang2018unreasonable},  VFID\cite{balaji2019conditional}, FVD\cite{unterthiner2018towards}. These metrics jointly assess both pixel-level reconstruction quality and perceptual-level fidelity, and are critical for evaluating the visual quality of human body generation in our task. Notably, we have introduced an additional quantitative metric in the benchmark: Percentage of Correct Keypoints (PCK)\cite{bergman2023generativeneuralarticulatedradiance}. This metric calculates the percentage of 2D keypoints detected on a generated frame that fall within a specified error threshold relative to the ground truth (conditioning) keypoints. It provides a more direct evaluation of pose structural fidelity in the generated results. 

In addition, to evaluate the value of our data set and the validity of our approach in our method, we introduce the three dimensions: Background consistency, overall consistency, smooth motion of Vbench-I2V\cite{huang2023vbench} for the quantitative evaluation of no real video participation with our base model Wan2.1-FunV1.1-Control\cite{Bub2025videoX} I2V.%it currently only supports the input of the first frame and pose-guided generation.

\subsection{Quantitative Results}
We compare our method with recent state-of-the-art baselines Animate-X \cite{tan2024animate}, ControlNeX \cite{peng2024controlnext}, StableAnimator \cite{tu2024stableanimator}, UniAnimate-DiT \cite{wang2025UniAnimate}, Wan2.1\text{-}Fun\text{-}V1.1\cite{Bub2025videoX}, on the HyperMotionX Bench. Uniformly use the pose video provided in our bench as the input condition. As shown in Table~\ref{tab:HYbenchmark_results}, our model achieves either the best or second-best performance across all metrics. Notably, our method delivers significant gains on pixel-level fidelity metrics (PSNR, SSIM, L1). Moreover, on structural accuracy, measured by PCK, our approach also achieves SOTA on the HyperMotionX bench with a substantial margin over all baselines. These results indicate that the proposed \textbf{SLF-RoPE} effectively enhances spatial structure while preserving low-frequency appearance in complex-motion frames.   %尤其是在结构性指标PCKh
On perceptual and temporal metrics (LPIPS, FID, VFID, FVD) our method remains highly competitive and consistently ranks among the top performers.
Taken together, the strong performance across low-level and high-level metrics demonstrates the effectiveness of our design for challenging human motion scenarios.

\begin{table*}[t]
    \centering
    \scriptsize   % or \tiny
    \caption{Quantitative comparison with state-of-the-art methods on the HyperMotionX Bench.
    $\uparrow$ indicates higher is better; $\downarrow$ indicates lower is better.
    \textcolor{red!80!black}{Red} and \textcolor{RoyalBlue}{Blue} denote the best and second-best, respectively.}
    \label{tab:HYbenchmark_results}

    \setlength{\tabcolsep}{4.0pt}
    \renewcommand{\arraystretch}{1.2}
    {
    \footnotesize
    \begin{tabular*}{\textwidth}{@{\extracolsep{\fill}}lcccccccc}
        \toprule
        \textbf{Method} &
        \textbf{PSNR}$\uparrow$ &
        \textbf{SSIM}$\uparrow$ &
        \textbf{L1}$\downarrow$ &
        \textbf{LPIPS}$\downarrow$ &
        \textbf{FID}$\downarrow$ &
        \textbf{VFID}$\downarrow$ &
        \textbf{FVD}$\downarrow$ &
        \textbf{PCK@0.5(\%)}$\uparrow$ \\
        \midrule
        Animate\text{-}X\cite{tan2024animate} &
        19.19 & 0.62 & $61.0 \times 10^{-3}$ & 0.210 & 110.32 & 17.07 & 1798.89 & 40.56 \\
        ControlNeXt\cite{peng2024controlnext} &
        20.39 & 0.64 & $58.9 \times 10^{-3}$ & 0.158 & 94.66 & 16.62 & 985.00 & 49.58 \\
        StableAnimator\cite{tu2024stableanimator} &
        19.82 & 0.63 & $62.9 \times 10^{-3}$ & 0.163 & 93.22 & 33.13 & 1184.88 & 17.84 \\
        UniAnimate\text{-}DiT &
        \textcolor{RoyalBlue}{20.90} &
        \textcolor{RoyalBlue}{0.68} &
        \textcolor{RoyalBlue}{$56.9 \times 10^{-3}$} &
        0.152 &
        \textcolor{red!80!black}{80.90} &
        \textcolor{red!80!black}{14.50} &
        981.43 &
        23.17 \\
        Wan2.1\text{-}Fun\text{-}V1.1\cite{Bub2025videoX} &
        20.72 &
        0.63 &
        $58.0 \times 10^{-3}$ &
        \textcolor{RoyalBlue}{0.148} &
        81.86 &
        17.89 &
        \textcolor{red!80!black}{801.74} &
        \textcolor{RoyalBlue}{65.62} \\
        \midrule
        \textbf{Hypermotion} &
        \textcolor{red!80!black}{22.03} &
        \textcolor{red!80!black}{0.71} &
        \textcolor{red!80!black}{$47.0 \times 10^{-3}$} &
        \textcolor{red!80!black}{0.124} &
        \textcolor{RoyalBlue}{80.91} &
        \textcolor{RoyalBlue}{16.49} &
        \textcolor{RoyalBlue}{825.68} &
        \textcolor{red!80!black}{70.32} \\
        \bottomrule
    \end{tabular*}
    }
\end{table*}

\begin{figure}[t]
    \centering
    \includegraphics[width=\linewidth]{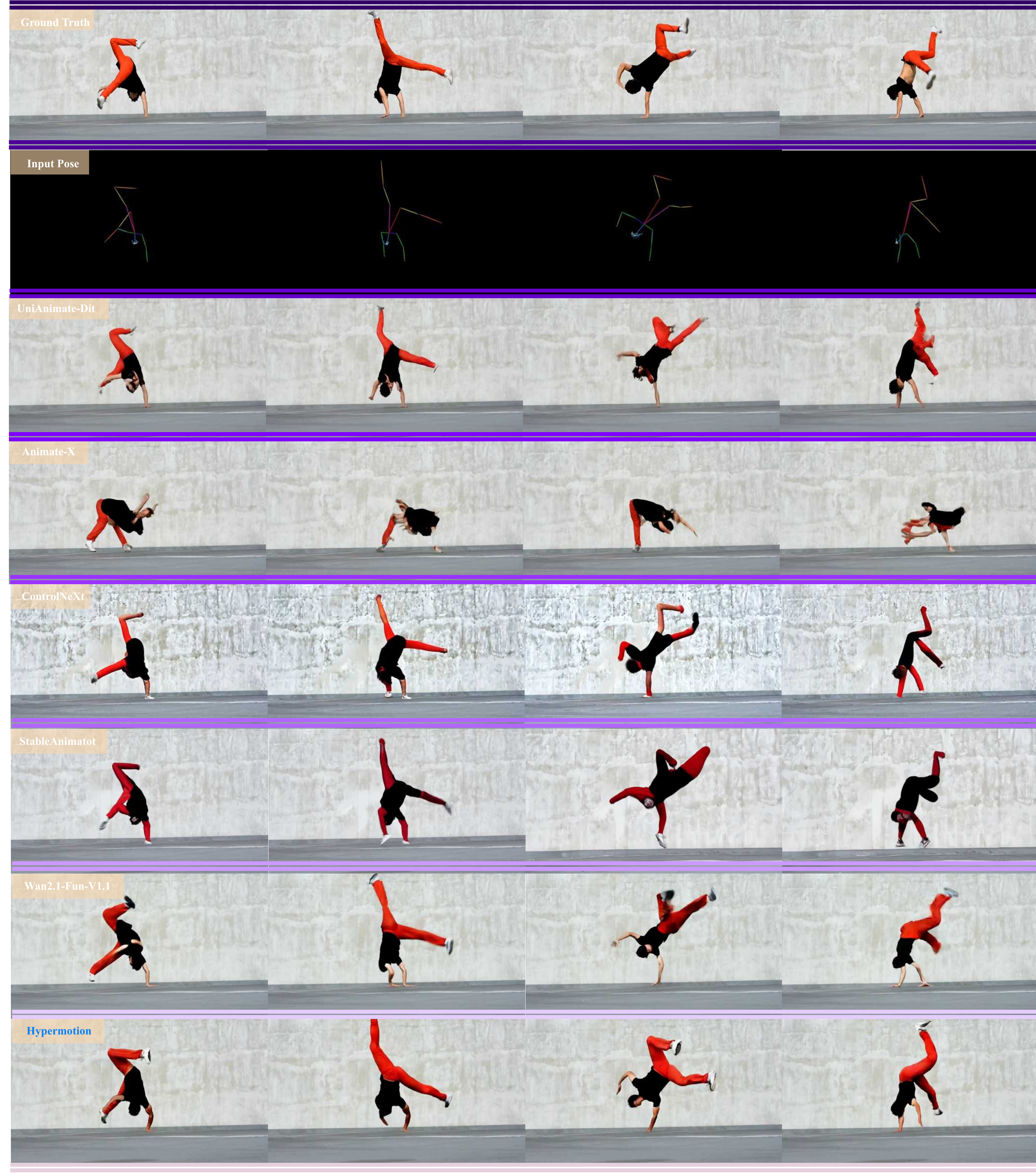}
    \caption{\small Qualitative comparison case \textit{a}. Break dance 2000s.}
    \label{fig:q1}
\end{figure}

\begin{figure}[t]
    \centering
    \includegraphics[width=\linewidth]{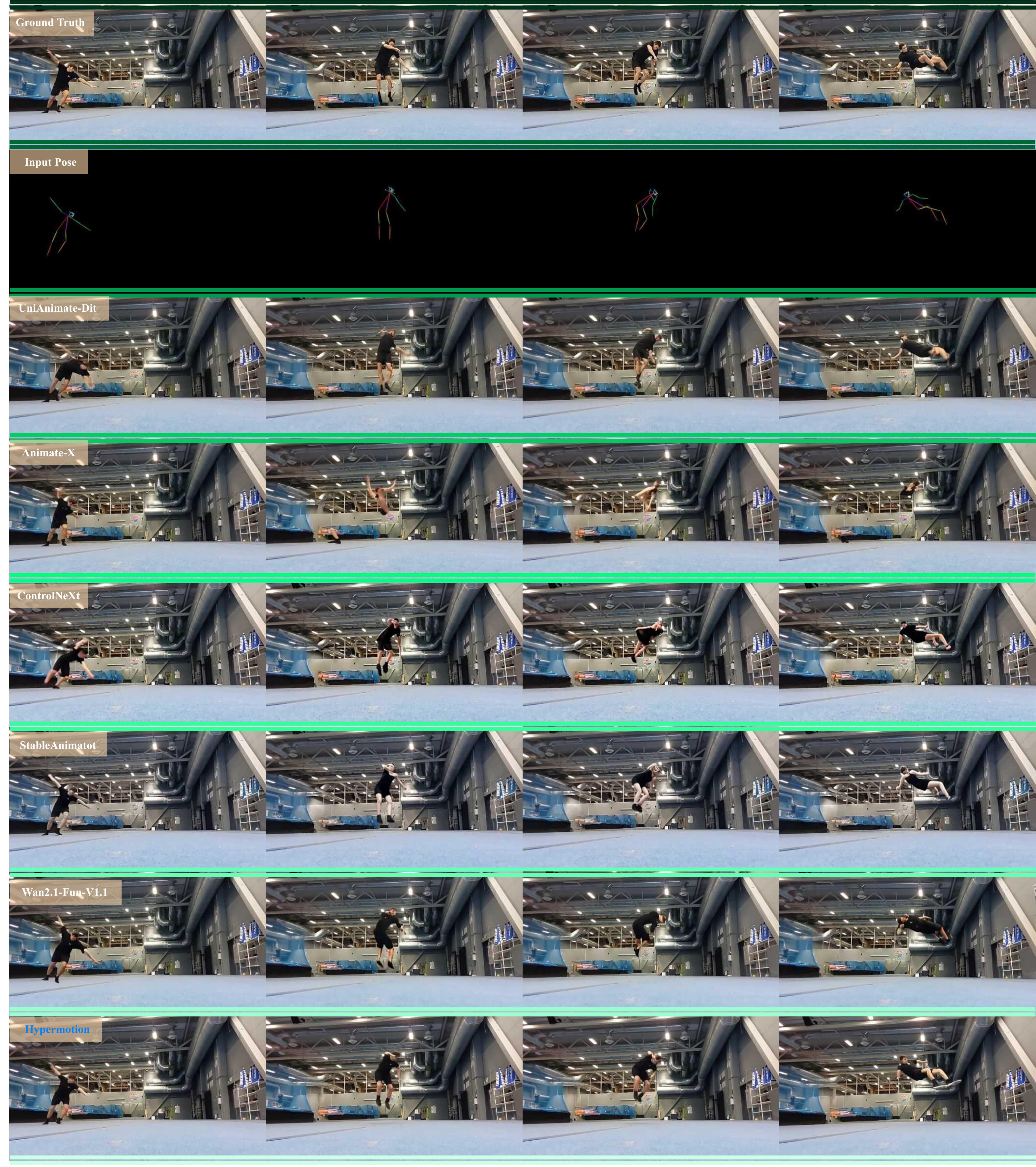}
    \caption{\small Qualitative comparison case \textit{b}. Front somersault.}
    \label{fig:q2}
\end{figure}

\begin{table}[t]
    \centering
    \caption{Ablation study on the effectiveness of the proposed Open-HyperMotion dataset.
    We evaluate the performance using VBench-I2V metrics\cite{huang2023vbench}.
    $\uparrow$ indicates higher is better.}
    \label{tab:Vbench_results}
    
    \renewcommand{\arraystretch}{1.30}
    \begin{tabular*}{\columnwidth}{@{\extracolsep{\fill}}lccc}
        \toprule
        \textbf{Method} &
        \makecell{\textbf{Background}\\\textbf{Consistency} $\uparrow$} &
        \makecell{\textbf{Overall}\\\textbf{Consistency} $\uparrow$} &
        \makecell{\textbf{Motion}\\\textbf{Smoothness} $\uparrow$} \\
        \midrule
        \makecell[l]{Wan2.1-Fun-Control\\I2V (Untrained)} & 94.14 & 10.57 & 98.99 \\
        \makecell[l]{Wan2.1-Fun-Control\\I2V (Trained)}   & 94.61 & 10.61 & 99.08 \\
        \midrule
        \textbf{Hypermotion (Ours)} & 95.00 & 10.54 & 99.19 \\
        \bottomrule
    \end{tabular*}
\end{table}

To further validate, We additionally adopt \textbf{VBench-I2V} as a no-reference evaluation protocol. We compare with our base model Wan2.1-Fun-V1.1-Control (I2V mode).
As shown in Table~\ref{tab:Vbench_results}, our method achieves superior performance across all three evaluation metrics: \textit{Background Consistency}, \textit{Overall Consistency}, and \textit{Motion Smoothness}.
These results further demonstrate the effectiveness of both our modeling framework 
and the proposed dataset in handling complex motion generation scenarios.
\subsection{Qualitative Results}
\begin{figure*}[t]
  \centering
  \includegraphics[width=0.7\textwidth]{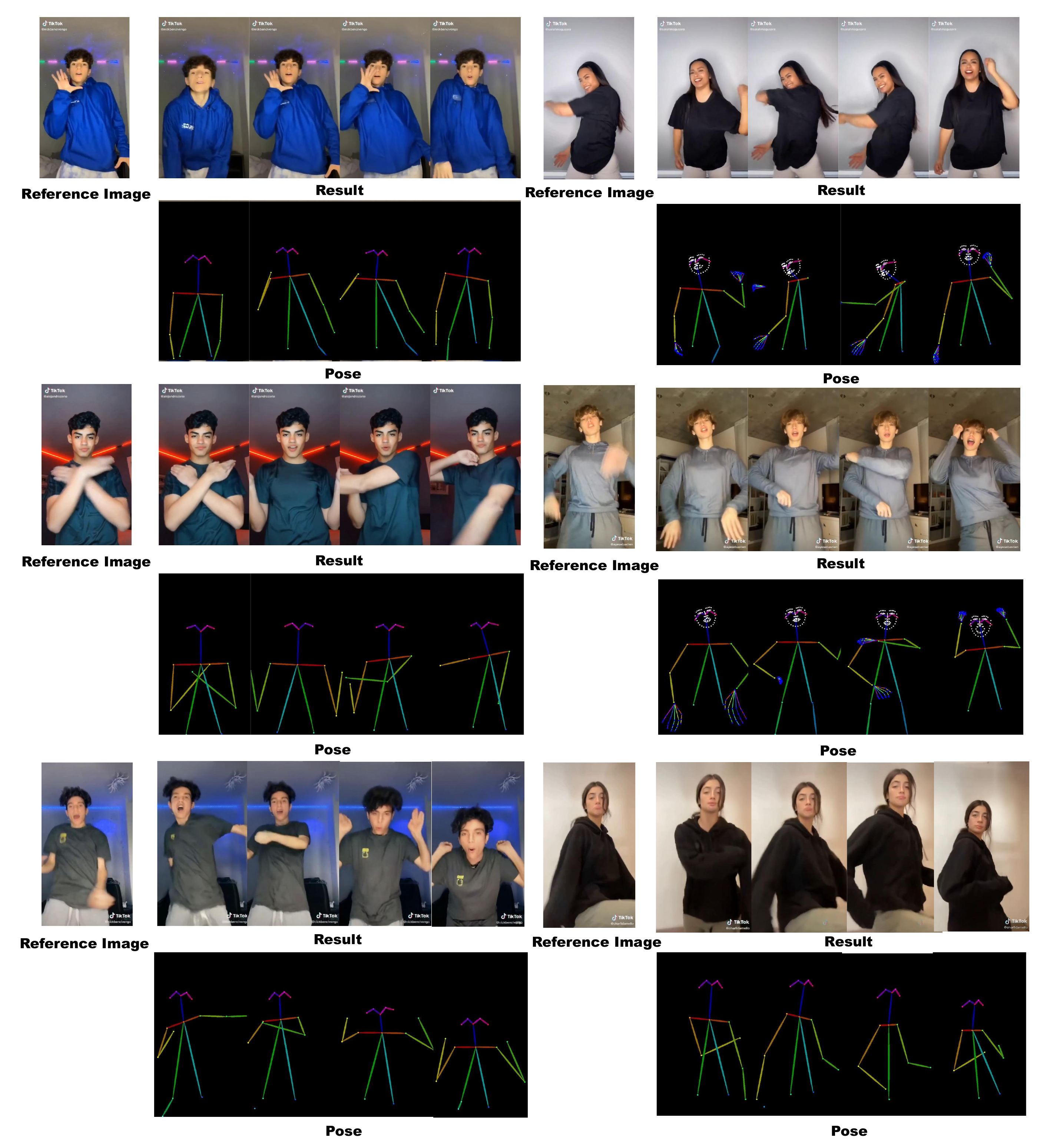}
  \caption{Further case studies highlighting reliable hand synthesis and consistent identity preservation in challenging high-motion scenes.}
  \label{fig:ID}
\end{figure*}

In Fig.~\ref{fig:D}, we present qualitative comparisons with recent state-of-the-art baselines on complex human motion scenarios from the HyperMotionX Bench. Additional qualitative results are provided in Fig.~\ref{fig:q1} and Fig.~\ref{fig:q2},~\ref{fig:ID}.

From the qualitative comparisons, several clear observations can be made. Compared with non-DiT-based methods, our approach demonstrates a significant advantage in terms of overall human body plausibility. 
Baseline methods without DiT architectures frequently suffer from severe body distortions, including unnatural limb bending, incorrect joint connectivity, and inconsistent body proportions, especially under fast or large-amplitude motions. In contrast, our method is able to generate structurally coherent human poses with well-preserved kinematic constraints, effectively mitigating deformation artifacts across the entire motion sequence. Second compared with DiT-based methods, our approach further exhibits superior accuracy and stability in challenging motion phases, such as airborne moments, rapid rotations, and complex transitions involving strong temporal dynamics. 
While existing DiT-based baselines tend to produce blurred poses, pose drifting, or local inconsistency during these high-difficulty moments, our method maintains precise body alignment and temporally consistent motion trajectories. 
This advantage is particularly evident in scenarios involving jumping, spinning, and abrupt direction changes. In addition, empirical results on the TikTok\cite{Jafarian_2021_CVPR} dataset demonstrate that our method achieves robust identity preservation and faithful hand motion generation even in common dance scenarios, as shown in Fig~\ref{fig:ID}.

Overall, these qualitative results indicate that our method not only improves basic human structure correctness over non-DiT approaches but also enhances fine-grained motion fidelity and temporal coherence beyond existing DiT-based video generation models, especially in highly dynamic and complex human motion scenarios.

\begin{figure*}[t]
  \centering
  \includegraphics[width=\textwidth]{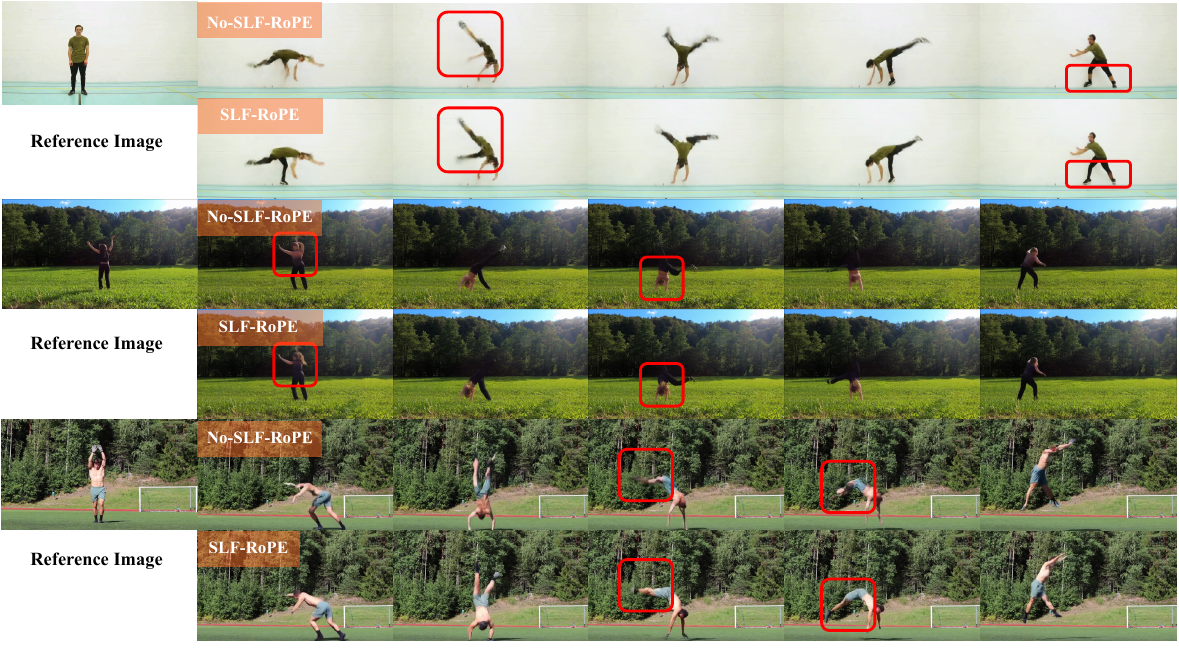}
  \caption{Ablation study on the impact of SLF-RoPE on the HyperMotionX Bench.
  The experimental control group results shown here are all from our models.}
  \label{fig:abl}
\end{figure*}

\begin{table*}
    \centering
    \caption{Ablation study on the impact of SLF-RoPE and the proposed Open-HyperMotion dataset on the HyperMotionX Bench.
    $\uparrow$ indicates higher is better; $\downarrow$ indicates lower is better.}
    \label{tab:HY_ablation_results}

    \setlength{\tabcolsep}{5.6pt}
    \renewcommand{\arraystretch}{1.12}
    \scriptsize

    \resizebox{\textwidth}{!}{%
    \begin{tabular}{lcccccccc}
        \toprule
        \textbf{Method} &
        \textbf{PSNR}$\uparrow$ &
        \textbf{SSIM}$\uparrow$ &
        \textbf{L1}$\downarrow$ &
        \textbf{LPIPS}$\downarrow$ &
        \textbf{FID}$\downarrow$ &
        \textbf{VFID}$\downarrow$ &
        \textbf{FVD}$\downarrow$ &
        \textbf{PCK@0.5(\%)}$\uparrow$ \\
        \midrule
        Baseline (Untrained) &
        20.72 & 0.63 & $58.0 \times 10^{-3}$ & 0.148 & 81.86 & 17.89 & 801.73 & 65.62 \\
        \addlinespace[2pt]
        Baseline (Trained) &
        21.78 & 0.69 & $48.3 \times 10^{-3}$ & 0.125 & 77.00 & 13.19 & 758.71 & 68.10 \\
        \addlinespace[2pt]
        \rowcolor{gray!10}
        \textbf{SLF-RoPE} &
        \textbf{22.03} &
        \textbf{0.71} &
        \textbf{$47.0 \times 10^{-3}$} &
        \textbf{0.124} &
        80.91 &
        16.49 &
        825.68 &
        \textbf{70.32} \\
        \bottomrule
    \end{tabular}%
    }
\end{table*}

\begin{table*}[h]
\centering
\caption{Ablation study on \textbf{TikTok} and \textbf{UBC-Fashion} benchmarks regarding the impact of SLF-RoPE. $\uparrow$ indicates higher is better; $\downarrow$ indicates lower is better.}
\label{tab:tiktok_fashion_ablation}
\setlength{\tabcolsep}{17.6pt}
\footnotesize
\begin{tabular}{lcccccc}
\toprule
Model & PSNR $\uparrow$ & SSIM $\uparrow$ & L1 $\downarrow$ & LPIPS $\downarrow$ & FID $\downarrow$ & FVD $\downarrow$ \\
\midrule
Remove SLF-RoPE (TikTok) & 16.98 & 0.64 & $89.7 \times 10^{-3}$ & 0.23 & 98.45 & 833.42 \\
\rowcolor{gray!10}
SLF-RoPE (TikTok)        & 16.52 & 0.63 & $97.4 \times 10^{-3}$ & 0.26 & 102.74 & 887.32 \\
\addlinespace[2pt]
Remove SLF-RoPE (Fashion) & 23.73 & 0.90 & $26.1 \times 10^{-3}$ & 0.052 & 29.80 & 172.47 \\
\rowcolor{gray!10}
SLF-RoPE (Fashion)        & 23.86 & 0.90 & $26.0 \times 10^{-3}$ & 0.059 & 38.31 & 243.29 \\
\bottomrule
\end{tabular}

\end{table*}

\subsection{Ablation Study}
Ablation on \textbf{SLF-RoPE}. To analyse the contribution of SLF-RoPE we removed SLF-RoPE from our baseline and compared it to our baseline with the same number of training steps. As shown in Table~\ref{tab:HY_ablation_results}, SLF-RoPE brings significant improvements to the model. For the Hypermotion model, SLF-RoPE contributes to improvements on pixel fidelity metrics (PSNR, SSIM, L1). On perceptual metrics (LPIPS, FID, VFID, FVD), we also observe improvements, albeit more modest than those over the baseline trained solely on our dataset. The slight decrease in perceptual performance is due to the perception–distortion (perceptual–fidelity) trade-off, rather than ineffectiveness of SLF-RoPE\cite{DBLP:journals/corr/abs-1711-06077,wang2025traversingdistortionperceptiontradeoffusing}. By contrast, SLF-RoPE delivers a clear gain in structural alignment, as evidenced by higher PCK score.

\begin{table*}[t]
\centering
\caption{Comparison of EasyAnimate5-7B\cite{xu2024easyanimatehighperformancelongvideo} (control-task) before and after training on Open-HyperMotionX, evaluated on the HyperMotionX Benchmark. $\uparrow$ indicates higher is better; $\downarrow$ indicates lower is better.}
\label{tab:easyanimate_before_after}

\small
\renewcommand{\arraystretch}{1.15}

\begin{tabular*}{\textwidth}{@{\extracolsep{\fill}}lcccccc}
\toprule
\textbf{Method} & \textbf{PSNR}$\uparrow$ & \textbf{SSIM}$\uparrow$ & \textbf{L1}$\downarrow$ & \textbf{LPIPS}$\downarrow$ & \textbf{FID}$\downarrow$ & \textbf{FVD}$\downarrow$ \\
\midrule
Untrained & 19.31 & 0.64 & $63.9 \times 10^{-3}$ & 0.196 & 111.09 & 1377.32 \\
Trained   & 19.74 & 0.66 & $59.4 \times 10^{-3}$ & 0.188 & 97.44  & 1129.42 \\
\bottomrule
\end{tabular*}
\end{table*}

\begin{table*}[t]
\centering
\caption{SLF-RoPE ablation study under low-quality pose (DWpose\cite{yang2023effective}) inputs on the HyperMotionX Bench. $\uparrow$ indicates higher is better; $\downarrow$ indicates lower is better.}
\label{tab:dwpose_ablation_results}

\small
\renewcommand{\arraystretch}{1.15}

\begin{tabular*}{\textwidth}{@{\extracolsep{\fill}}lcccccc}
\toprule
\textbf{Method} & \textbf{PSNR}$\uparrow$ & \textbf{SSIM}$\uparrow$ & \textbf{L1}$\downarrow$ & \textbf{LPIPS}$\downarrow$ & \textbf{FID}$\downarrow$ & \textbf{FVD}$\downarrow$ \\
\midrule
Remove SLF-RoPE & 20.65 & 0.63 & $60.3 \times 10^{-3}$ & 0.1523 & 81.04 & 833.58 \\
SLF-RoPE        & 21.26 & 0.67 & $52.1 \times 10^{-3}$ & 0.1366 & 83.03 & 882.51 \\
\bottomrule
\end{tabular*}
\end{table*}

\begin{table*}[t]
\centering
\caption{SLF-RoPE ablation study using different frequency-domain enhancement strategies trained on the HyperMotionX Bench. $\uparrow$ indicates higher is better; $\downarrow$ indicates lower is better.}
\label{tab:slf_rope_freq_ablation}

\small
\renewcommand{\arraystretch}{1.15}

\begin{tabular*}{\textwidth}{@{\extracolsep{\fill}}lcccccc}
\toprule
\textbf{Model} &
\textbf{PSNR}$\uparrow$ &
\textbf{SSIM}$\uparrow$ &
\textbf{L1}$\downarrow$ $(\times 10^{-3})$ &
\textbf{LPIPS}$\downarrow$ &
\textbf{FID}$\downarrow$ &
\textbf{FVD}$\downarrow$ \\
\midrule
No Enhance & 21.78 & 0.69 & 48.3 & 0.1255 & 77.00 & 758.71 \\
Enhance H  & 21.14 & 0.68 & 52.4 & 0.1409 & 85.68 & 827.27 \\
Enhance L  & 22.03 & 0.71 & 47.0 & 0.1240 & 80.91 & 825.68 \\
\bottomrule
\end{tabular*}
\end{table*}

Visualization of the ablation study in SLF-RoPE on the HyperMotionX Bench, as shown in Fig~\ref{fig:abl}. In the first and second cases, we highlight the regions with red zoom-in boxes for detailed comparison. We observe that removing SLF-RoPE and relying solely on our baseline leads to inconsistencies between the generated human appearance and the input reference image. In the first case, the person in the reference image is wearing long black trousers, but the baseline model without SLF-RoPE instead generates short black pants. The discrepancy is even more pronounced in the second case, where the model fails to accurately reproduce the black T-shirt from the reference image.

The third case further reveals that removal of SLF-RoPE results in severe degradation during airborne motion frames: specifically, the lower limbs of the character become incomplete or distorted, making it difficult to recognize the leg structure. In contrast, the generation results with SLF-RoPE retain the correct appearance and structure, effectively mitigating such artifacts.

We further evaluate the impact of SLF-RoPE on general-motion benchmarks, including TikTok\cite{Jafarian_2021_CVPR} and UBC-Fashion\cite{Zablotskaia2019DwNet}. As shown in the Table\textbf{~\ref{tab:tiktok_fashion_ablation}}, the results indicate that SLF-RoPE does not degrade the performance in these common motion datasets, achieving fidelity and perceptual quality comparable to baseline. This confirms that our frequency aware modification remains stable on simple motions while primarily benefiting complex motion scenarios.

In addition, we further analyze the impact of training on the Open-HyperMotionX dataset by evaluating the EasyAnimate5-7B\cite{xu2024easyanimatehighperformancelongvideo} model (control-task) before and after finetuning, using the HyperMotionX Benchmark as a testbed. As shown in Table~\ref{tab:easyanimate_before_after}, finetuning leads to consistent improvements in most metrics, including PSNR, SSIM, L1, LPIPS, FID, and FVD, indicating enhanced visual quality, temporal coherence, and pixel-level fidelity. This shows that the proposed Open-HyperMotionX dataset effectively increases the ability of a different baselines to model complex human motions animation.

The effect of SLF-RoPE under Low Quality Pose, as show in Table~\ref{tab:dwpose_ablation_results}.
To assess the robustness of SLF-RoPE, we replace high-quality pose inputs with degraded poses generated by DWpose\cite{yang2023effective}. Table~\ref{tab:dwpose_ablation_results} shows that incorporating SLF-RoPE still improves PSNR, SSIM, and L1, demonstrating that the module remains effective even when pose guidance is partially missing or noisy. This aligns with our motivation: SLF-RoPE leverages the diffusion prior to compensate for unreliable structural input, yielding more stable and coherent results.

Finally, we compare different variants of frequency-domain enhancement within SLF-RoPE. As shown in Table~\ref{tab:slf_rope_freq_ablation}, enhancing the low-frequency band (Enhance L) yields the best overall performance across reconstruction and perceptual metrics, confirming that low-frequency positional stability plays a dominant role in preserving large-scale spatial structure under complex human motion. In contrast, enhancing high-frequency signals (Enhance H) leads to degradation, further validating our design choice. It should be noted that during training, we employed identical training steps and hyperparameter settings.

\section{Conclusion}
\label{sec: Conclusion}
In this work, we propose a simple yet effective DiT-based baseline and introduced the \textbf{SLF-RoPE} module to enhance spatial low-frequency modeling for pose-guided human image animation under complex motion conditions. We presented the Open-HyperMotionX Dataset and HyperMotionX Bench to establish a new benchmark for evaluating this challenging task.  
Extensive experiments on these benchmarks demonstrate that our method consistently outperforms existing SOTA approaches across both pixel-level metrics (PSNR, SSIM, L1), structural metric (PCK) and perceptual metrics (LPIPS, FVD, FID, VFID). Notably, gains in pixel-level accuracy and structural alignment validate that our proposed SLF-RoPE better enhances global appearance and structure.   
Qualitative results further confirm superior consistency of identity, accuracy of the limb, and motion stability in various scenarios. We hope our dataset, benchmark, and method will benefit the research community.

\subsection{Limitations}
While our method demonstrates strong performance in a wide range of complex human motion scenarios, several limitations remain.  
First, the model still struggles with extremely dynamic and acrobatic movements such as continuous Thomas flares, 540 kicks, and other multi-phase actions. These motions often involve rapid full-body rotations and self-occlusions, which are inherently difficult to reconstruct with high fidelity. Second, under such extreme motions, facial features are often blurred or partially lost, indicating a lack of robustness in preserving fine-grained identity cues during high-speed transitions. Lastly, we observe that even state-of-the-art pose estimators fail to provide accurate hand keypoint sequences in these complex motion segments, making it challenging to guide fine-level motion generation in the hands and fingers.

We believe that future work can address these issues by incorporating explicit face and hand modeling.

% --- 参考文献 ---
% 注意：这里调用名为 references.bib 的文件
\bibliographystyle{IEEEtran}
\bibliography{reference}

\begin{IEEEbiography}[{\includegraphics[width=1in,height=1.25in,clip,keepaspectratio]{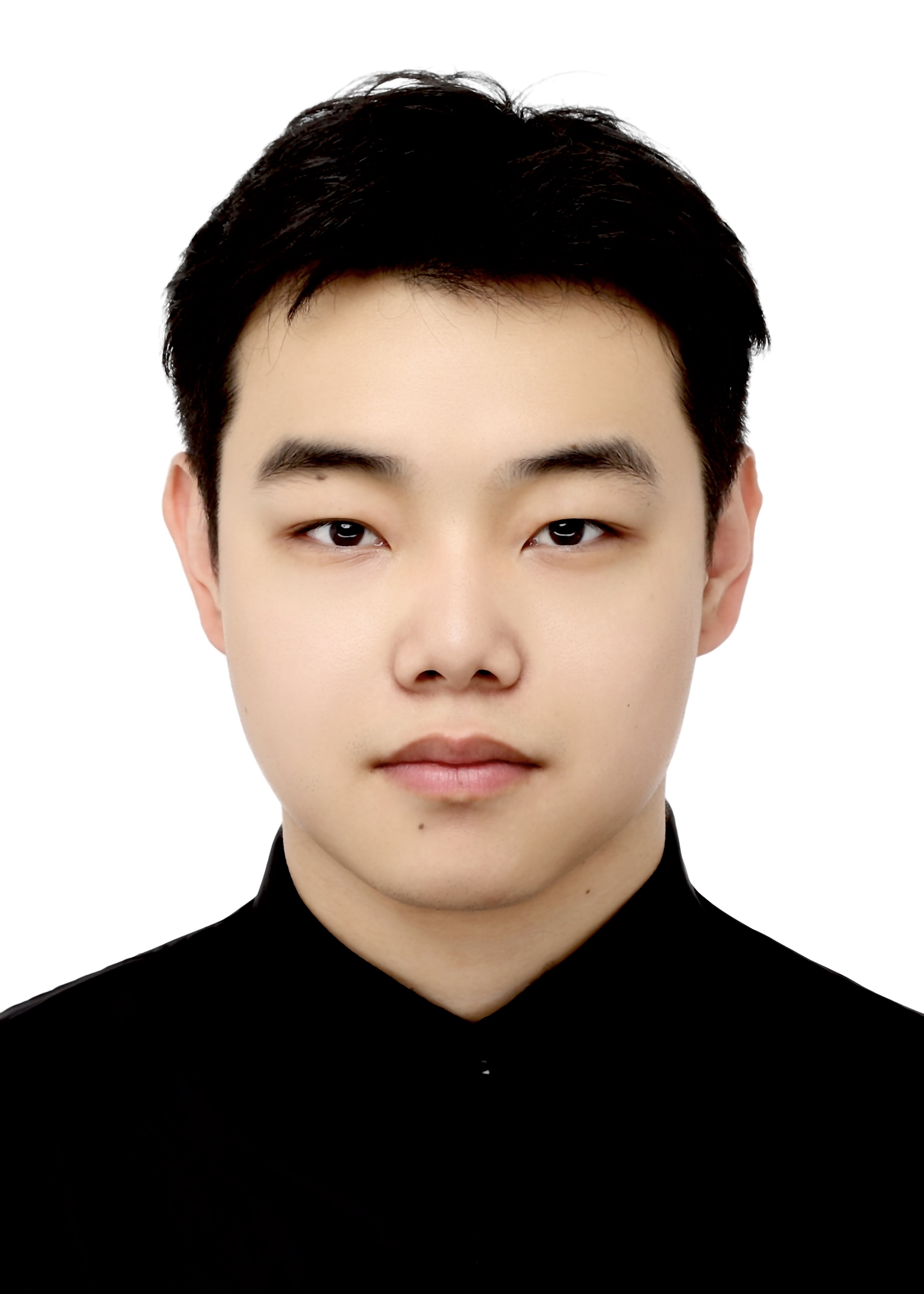}}]{Shuolin Xu}
Shuolin Xu is a Ph.D student at the National Centre for Computer Animation (NCCA) of Bournemouth University. He obtained his bachelor's degree from Zhengzhou University and his master's degree from Bournemouth University. Currently, he is conducting research on generative models, particularly in the areas of video generation and motion generation.
\end{IEEEbiography}

\begin{IEEEbiography}
[{\includegraphics[width=1in,height=1.25in,clip,keepaspectratio]{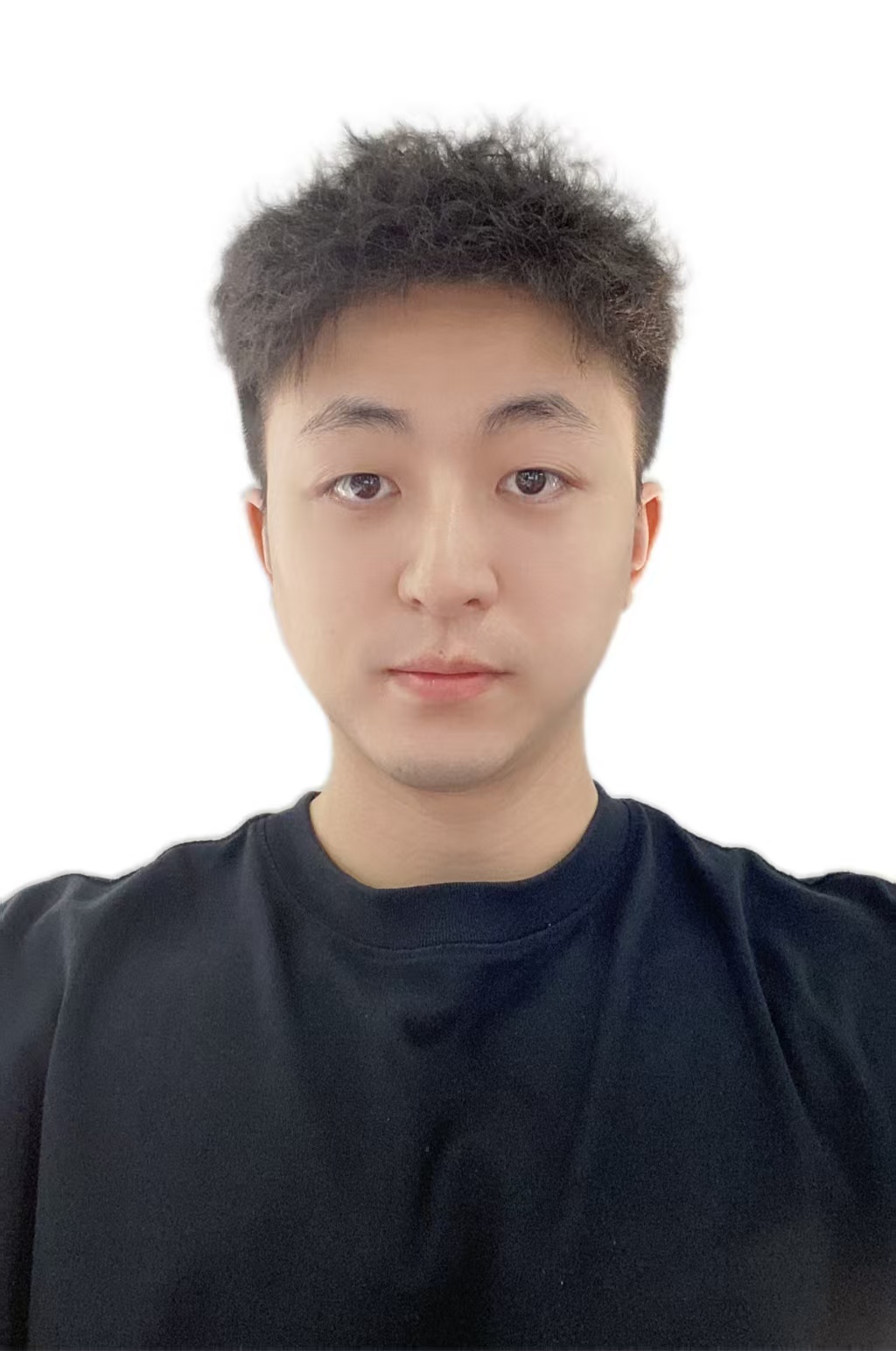}}]{Siming Zheng} received the B.S. degree from China Agricultural University and Ph.D degree from University of Chinese Academy of Sciences in 2023. Currently serves as researcher at vivo Mobile Communication Co., Ltd. His research interests include computer vision, generative model and computational imaging.
\end{IEEEbiography}

\begin{IEEEbiography}
[{\includegraphics[width=1in,height=1.25in,clip,keepaspectratio]{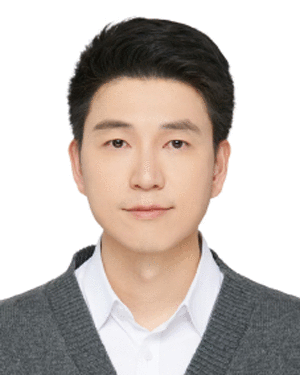}}]{Jinwei Chen} received the bachelor’s and PhD degrees from Zhejiang University. He is a member of company VIVO. His research interests include image processing and computational photography
\end{IEEEbiography}

\begin{IEEEbiography}
[{\includegraphics[width=1in,height=1.25in,clip,keepaspectratio]{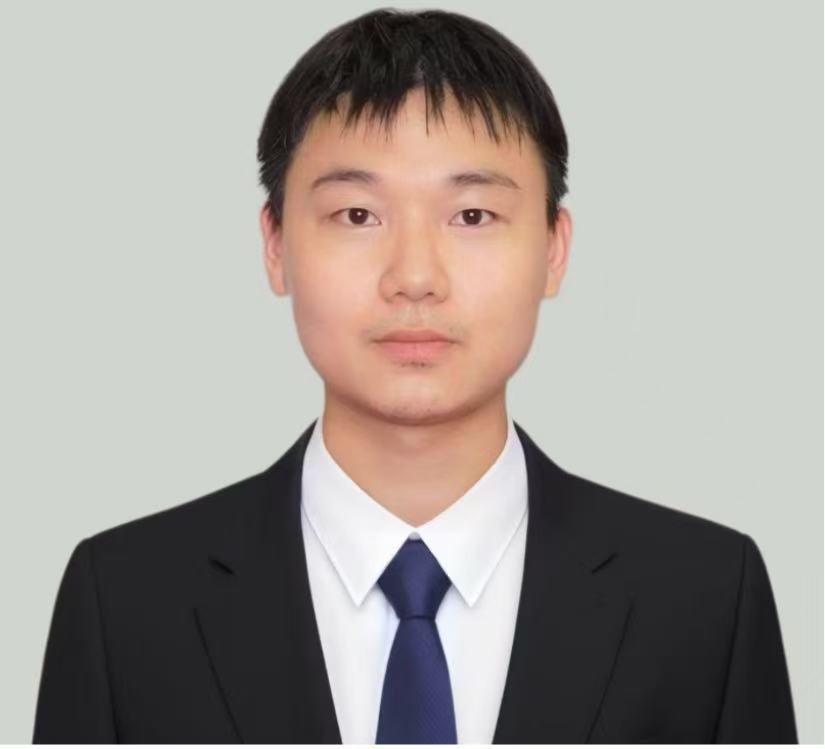}}]{Daqiang Zhou} Ph.D., is an Assistant Professor and Ph.D. Supervisor at both the School of Artificial Intelligence for Science and the School of Information Engineering, Peking University. He also serves as Assistant Dean of the School of Artificial Intelligence for Science and Head of the Preparatory Group for the Institute of Scientific Intelligence. He has been recognized among the World’s Top 2\% Scientists, selected for the National High-Level Young Talent Program, honored as an Outstanding Ph.D. Graduate in Singapore (awarded to ten individuals nationwide each year), and ranked among the Global Top 100 Talent Portraits. He is also a recipient of the Singapore President’s Science and Technology Award. His primary research interests include robotics and agent decision-making, efficient and hardware-friendly multimodal generation and understanding models, and edge computing.
\end{IEEEbiography}

\begin{IEEEbiography}[{\includegraphics[width=1in,height=1.25in,clip,keepaspectratio]{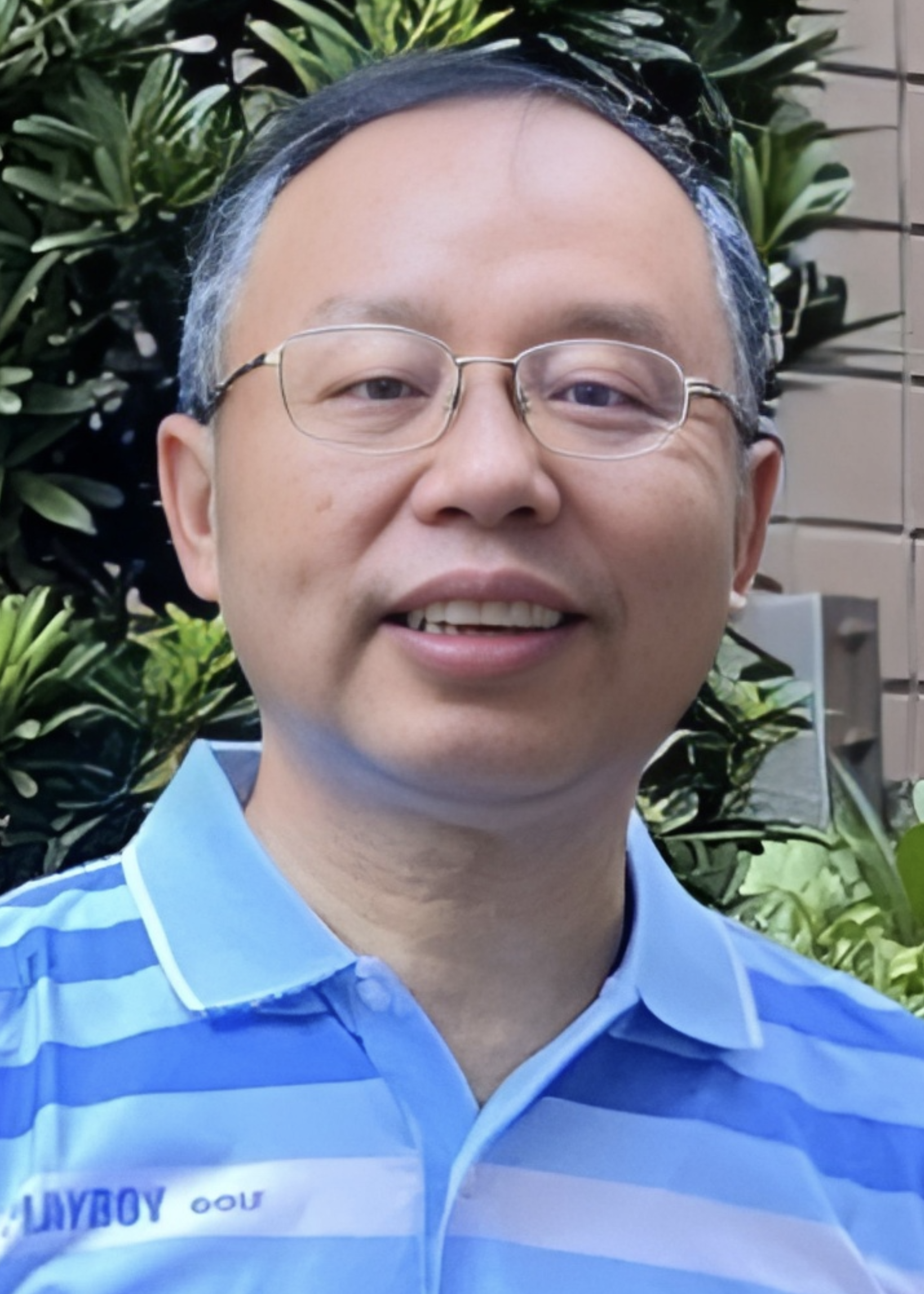}}]{Tong-Yee Lee}
Tong-Yee Lee (Senior Member, IEEE) received the Ph.D. degree in computer engineering from Washington State University, Pullman, in 1995. He is currently a chair professor with the Department of Computer Science and Information Engineering, National Cheng-Kung University (NCKU), Tainan, Taiwan. He leads the Computer Graphics Laboratory, NCKU (http://graphics.csie.ncku.edu.tw). His current research interests include computer graphics, nonphotorealistic rendering, medical visualization, virtual reality, and media resizing. He is a Senior Member of the IEEE and a Member of the ACM. He is an Associate Editor of the IEEE Transactions on Visualization and Computer Graphics. He also serves on the editorial boards of both the IEEE Transactions on Visualization and Computer Graphics, and IEEE Computer Graphics and Applications.
\end{IEEEbiography}

\begin{IEEEbiography}[{\includegraphics[width=1in,height=1.25in,clip,keepaspectratio]{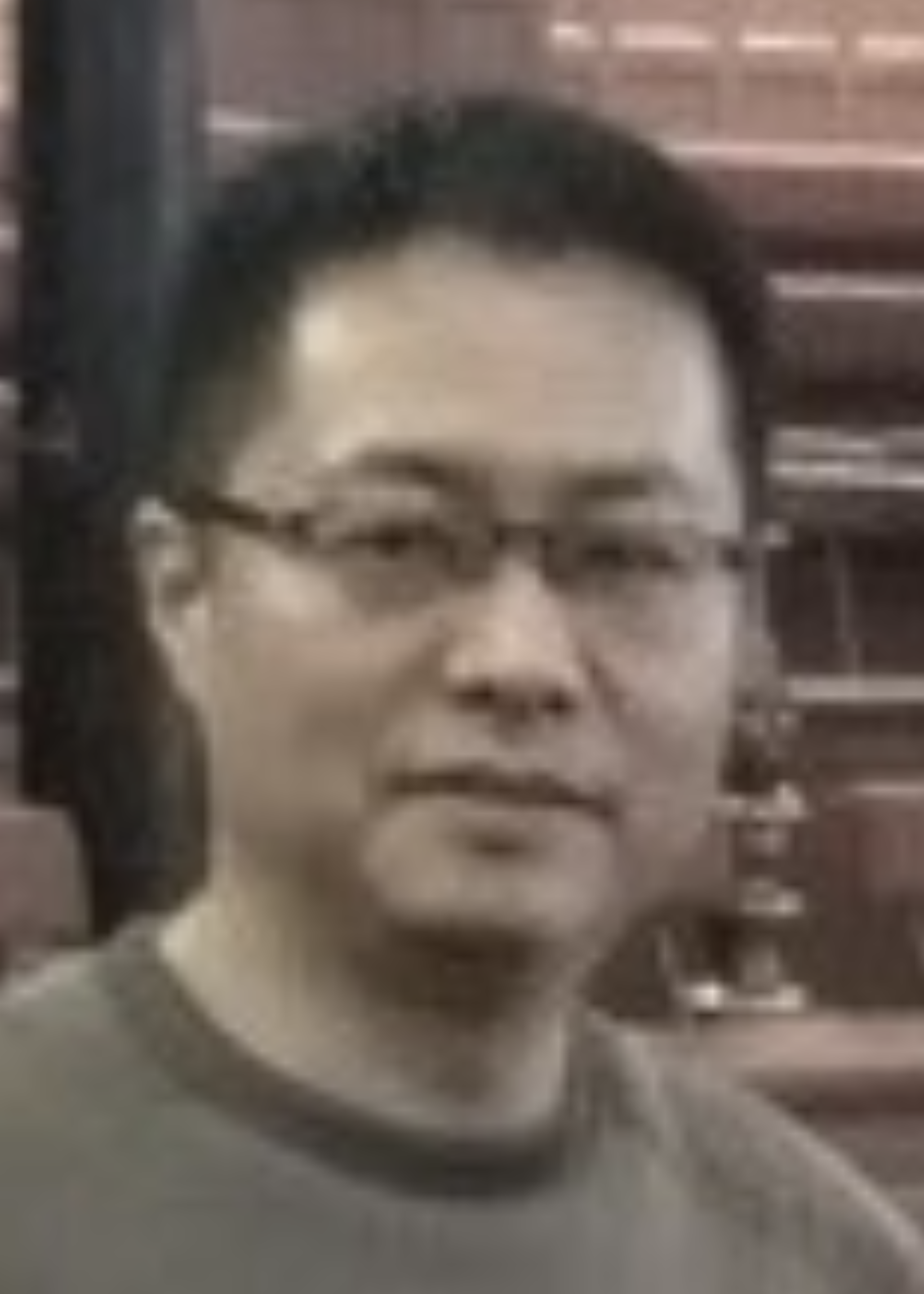}}]{Hongchuan Yu}
Hongchuan Yu is a Principal Academic of computer graphics in National Centre for Computer Animation, Bournemouth University, UK. He has published around 110 academic articles in reputable journals and conferences, and regularly served as PC members/referees for international journals and conferences. He is a Member of IEEE (MIEEE) and a fellow of High Education of Academy United Kingdom (FHEA). His research interests include Geometry, GenAI, Graphics, Image, and Video processing.
\end{IEEEbiography}

\begin{IEEEbiography}
[{\includegraphics[width=1in,height=1.25in,clip,keepaspectratio]{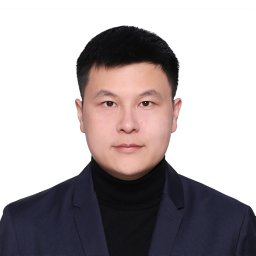}}]{Bo Li} (Member, IEEE) received the BSc and PhD
degrees from the Department of Computer Science, Nanjing University, China, in 2014 and 2019, respectively. From 2020 to 2023, he was a senior researcher with the Youtu Lab, Tencent, China. He is currently
a senior expert with the Vivo Image Algorithm Research Department, China. His research interests include computer vision, pattern recognition, and artificial intelligence.
\end{IEEEbiography}

\begin{IEEEbiography}
[{\includegraphics[width=1in,height=1.25in,clip,keepaspectratio]{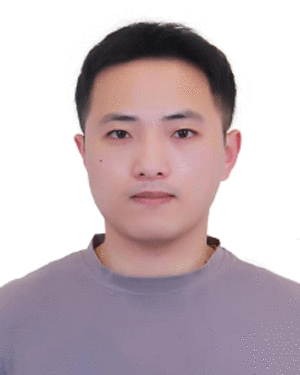}}]{Peng-Tao Jiang} received the PhD degree from
Nankai University, advised by Prof. Ming-MingCheng. He is currently a lead researcher engineer with the Quality Enhancement Center of vivo. Before that, he was a post-doc researcher with Zhejiang University, working with Prof. Chunhua Shen. His current research interests include diffusion, image restoration, multi-task learning, and segmentation.
\end{IEEEbiography}

\vfill

\end{document}